\UseRawInputEncoding
\documentclass[runningheads]{misc/llncs}

\usepackage{misc/eccv}

\usepackage{misc/eccvabbrv}

\usepackage{graphicx}
\usepackage{booktabs}
\usepackage{pgfplots}
\usepackage{subcaption}

\usepackage[accsupp]{axessibility}  

\usepackage{hyperref}
\usepackage{paralist}
\usepackage{array}
\usepackage{multirow}
\usepackage{amsmath}
\usepackage{colortbl}
\usepackage{soul}

\usepackage[framemethod=TikZ]{mdframed}
\usepackage{lipsum}

\def\ie{\emph{i.e.}\xspace}
\newcommand{\mypartitle}[2][2.25]{\vspace*{-#1 ex}~\\{\noindent {\bf #2}}}

\newcommand\sam[1]{{\color{black} #1}}

\usepackage{orcidlink}

\usepackage{pifont}

\newcommand{\cmark}{\ding{51}}%
\newcommand{\xmark}{\ding{55}}%
\definecolor{BackgroundColor}{RGB}{247, 217, 250}

\begin{document}

\title{ChildPlay-Hand: A Dataset of Hand Manipulations in the Wild}

\titlerunning{ChildPlay-Hand}

\author{Arya Farkhondeh* \and
Samy Tafasca* \and
Jean-Marc Odobez
}

\authorrunning{A. Farkhondeh et al.}

\institute{Idiap Research Institute, Martigny, Switzerland \\
\'Ecole Polytechnique F\'ed\'erale de Lausanne, Switzerland \\
\email{\{afarkhondeh, stafasca, odobez\}@idiap.ch}
}

\maketitle

\begin{figure}
\vspace{-7mm}
    \centering
    \captionsetup{type=figure}
    \includegraphics[width=1.0\textwidth]{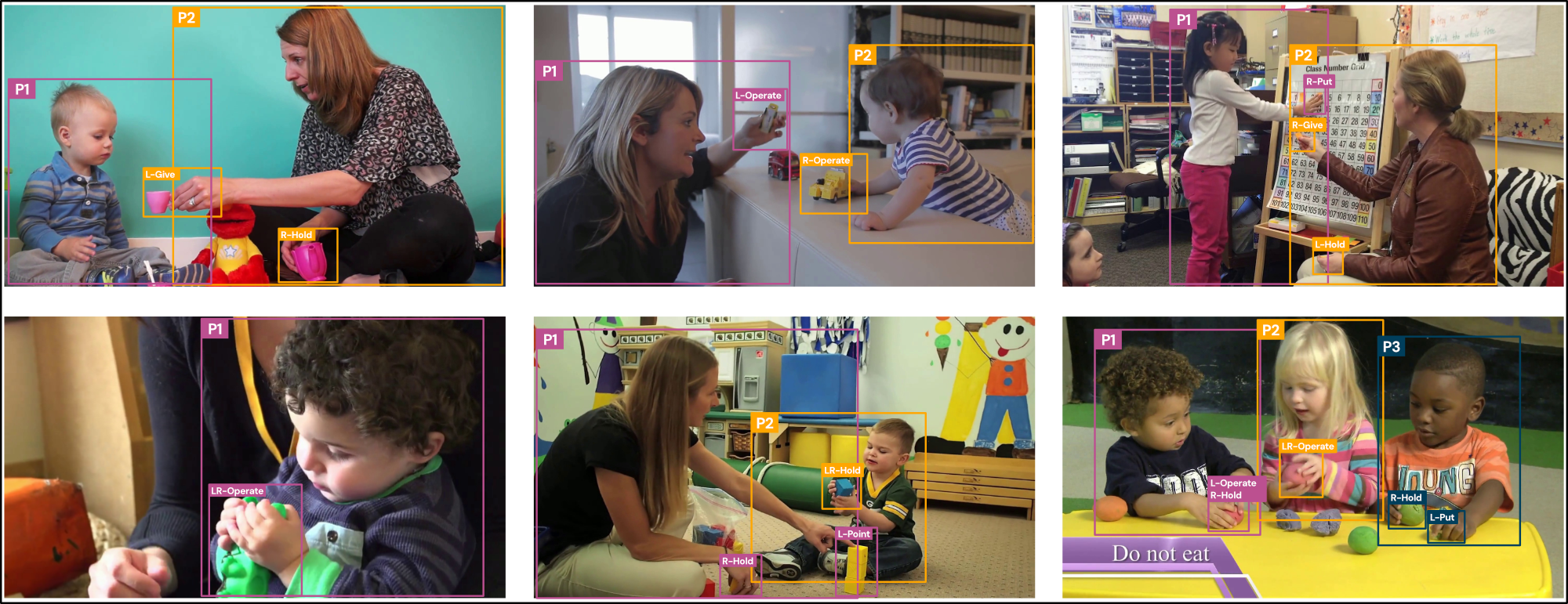}
    \vspace{-5mm}
    \captionof{figure}{Sample instances from the ChildPlay-Hand dataset with person bounding boxes and the per-hand object bounding boxes and corresponding action classes.}
\vspace{-15mm}
\label{fig:qualitative}
\end{figure}

\makeatletter
\renewcommand\@makefnmark{}
\makeatother
\footnotetext{* equal contribution}

\begin{abstract}

Hand-Object Interaction (HOI) is gaining significant attention, particularly with the creation of numerous egocentric datasets driven by AR/VR applications. However, third-person view HOI has received less attention, especially in terms of datasets. Most third-person view datasets are curated for action recognition tasks and feature pre-segmented clips of high-level daily activities, leaving a gap for in-the-wild datasets.
To address this gap, we propose ChildPlay-Hand, a novel dataset that includes person and object bounding boxes, as well as manipulation actions. ChildPlay-Hand is unique in: (1) providing per-hand annotations; 
(2) featuring videos in uncontrolled settings with natural interactions, involving both adults and children; (3) including gaze labels from the ChildPlay-Gaze dataset for joint modeling of manipulations and gaze. 
The manipulation actions cover the main stages of an HOI cycle, such as grasping, holding or operating, and different types of releasing. 
To illustrate the interest of the dataset, 
we study two tasks: 
object in hand detection (OiH), i.e. if a person has an object in their hand,
and manipulation stages (ManiS), which is more fine-grained and targets 
 the main stages of manipulation. We benchmark various spatio-temporal and segmentation networks, exploring body vs. hand-region information and comparing pose and RGB modalities. Our findings suggest that ChildPlay-Hand is a challenging new benchmark for modeling HOI in the wild.

 \vspace{-3mm}
\end{abstract}

\section{Introduction}
\label{sec:intro}

Hands are our  primary means of physical interaction with the environment, particularly through object manipulation. 
This explains why understanding hands in action has been an important topic in computer vision, focusing on various aspects such as recognizing gestures (see the recent review~\cite{Mohamed2021ARO}) and hand-object interaction (HOI)~\cite{FirstPersonAction_CVPR2018, Damen_2018_ECCV, Grauman_2022_CVPR, sener2022assembly101, Kwon_2021_ICCV}.
This understanding naturally finds applications in  various domains, including human-computer interaction~\cite{Rautaray2012VisionBH}, robotics~\cite{Yang2015RobotLM}, and augmented and virtual reality (AR/VR)~\cite{Hasson_2019_CVPR}.

Recently, much research has concentrated on HOI from a first-person perspective due to its applications in AR/VR. 
This has led to the creation of numerous egocentric video datasets, each tailored to specific settings and activities. These range from cooking activities in kitchen environments
(FPHA~\cite{FirstPersonAction_CVPR2018} and EPIC-Kitchen~\cite{Damen_2018_ECCV}), to more diverse scenarios in Ego4D~\cite{Grauman_2022_CVPR}.
Also, some other HOI datasets are captured from multiple views and are collected in lab environment and come with rich data like 3D hand pose and/or 6D object poses thanks to the use of  dedicated setup, as in H2O~\cite{Kwon_2021_ICCV} and Assembly101~\cite{sener2022assembly101}. 

Despite the extensive efforts in egocentric HOI, the study of hands from a third-person perspective has received less attention. Most third-person view datasets~\cite{soomro2012ucf101, kuehne2011hmdb, Heilbron2015ActivityNetAL, Kay2017TheKH, ntu_60, ntu_120, Gu_2018_CVPR, Sigurdsson2016HollywoodIH} are designed for action recognition tasks, where isolated short clips are annotated with human activities. These datasets primarily focus on high-level human activities in daily life, such as driving a car or washing dishes, serving a different research purpose. 
As a result, there is a notable lack of  datasets specifically concentrating on hands interacting with objects in the wild from a third-person view. 

To address this gap, we introduce ChildPlay-Hand. 
It is derived from the ChildPlay~\cite{tafasca2023childplay} (or ChildPlay-Gaze) video dataset, which was originally introduced for addressing the gaze following task for children and toddlers, and features videos from childcare facilities and school settings.
We selected this dataset because it contains both adults and children, 
and due to the rich diversity of hand manipulation behaviors mixed within 
children and adults interactions.
Building upon this dataset, we provide dense annotations of body bounding boxes, hand manipulation action for each hand,  and object bounding boxes of objects involved in such interactions. 
In terms of actions, we are interested in the main stages of the hand-object manipulation cycle: grasping, holding (passively having an object in hand) or operating (i.e. doing something actively with the object), and releasing (with several variants, see Sec.~\ref{sec:dataset}). 
While this may seem crude, this level of granularity allows to cover the entire hand-object interaction activities without being limited to the vocabulary of a particular application domain, and is anyway a first level of analysis that needs to be performed in HOI. 
Also, when addressed in the wild, it is actually very challenging. 

The above annotation scheme results in a dataset that is unique in several ways.
First, it provides \textbf{per-hand} annotations as hands can do multi-tasking. This is rare among existing datasets that typically focus on the activities of both hands as a whole. Second, it features videos in \textbf{uncontrolled settings} with varying camera views, showcasing scenes with multiple people interacting naturally with each other and objects in a free manner. This is in stark contrast with datasets that assume a fixed view with a single person performing a given activity, where hands and objects typically stay visible.
Third, the decision to use videos from ChildPlay-Gaze~\cite{tafasca2023childplay} will enable in the future the study of  the \textbf{coordination between manipulations and gaze} from a third-person view, which, to the best of our knowledge, has not been explored before. 
Lastly, \textbf{the inclusion of children} (and not dominantly adults) in the dataset makes it a valuable source for behavioral studies of this demographic group. Such datasets are typically private due to the sensitive nature of the data. 

ChildPlay-Hand can serve as a new benchmark for several tasks such as action recognition, action localization, and human-object interaction. 
In this work, we demonstrate its use and challenge for addressing two temporal action segmentation (TAS) tasks:
object in hand (OiH) and manipulation stages (ManiS). 
The former is a coarse task of predicting whether an object is in a given hand,  while the latter is more fine-grained, aiming to predict the main stages of hand-object manipulation interaction. For benchmarking, we explore different spatio-temporal networks like PoseConv3D~\cite{Duan_2022_CVPR}, RGBPoseConv3D~\cite{Duan_2022_CVPR}, and Hiera~\cite{ryali2023hiera} by fine-tuning them on ChildPlay-Hand. We also explore other aspects such as the use of hand inputs compared to full body inputs. We then use the best spatio-temporal network to extract features for full hand sequences as input to TAS methods such as MS-TCN~\cite{AbuFarha2019MSTCNMT} and report frame-based and segmental metrics. In summary, our \textbf{contributions} are as follows:
\begin{compactitem}
    \item We propose ChildPlay-Hand, a unique and novel dataset of hand 
    which provides per-hand activity annotations, features both adults and children, in uncontrolled and natural settings, and nicely complements existing gaze labels from ChildPlay-Gaze\cite{tafasca2023childplay}; 
    \item We benchmark the dataset for two tasks, investigating different state-of-the-art  spatio-temporal action recognition networks, as well as Temporal Action segmentation methods. 
     Our work establishes several baselines, comparing pose-only networks to multimodal and visual-only networks, and explores the effectiveness of full-body vs. hand region inputs.
\end{compactitem}

\label{sec:RelatedWork}
\section{Related Datasets}

\subsection{Hand-Object Interaction (HOI) datasets}

Recent video datasets of HOI are predominantly egocentric, each focusing on specific activities and settings. 
For instance, FPHA~\cite{FirstPersonAction_CVPR2018} and EPIC-Kitchen~\cite{Damen_2018_ECCV} capture cooking activities in kitchen environments, while Ego4D~\cite{Grauman_2022_CVPR} contains more diverse scenes, environments, and daily activities. 
In addition to egocentric view, some datasets offer multiple views, typically captured in controlled lab settings with dedicated setups, such as H2O~\cite{Kwon_2021_ICCV} and Assembly101~\cite{sener2022assembly101}. These datasets often include 3D hand poses and/or 6D object poses. 
In terms of activities, Assembly101~\cite{sener2022assembly101} contains fine-grained interaction verbs used to describe assembling and disassembling actions, such as "unscrew" and "remove" to achieve broader actions like "detach". Similarly, H2O~\cite{Kwon_2021_ICCV} defines 11 action verbs, including "open", "apply", and "read", to cover actions performed by participants when interacting with 8 distinct objects. 

Despite great contributions to egocentric HOI, there is a notable lack of third-person view datasets capturing hand manipulations in uncontrolled environments. 
We fill this gap with ChildPlay-Hand. While Assembly101~\cite{sener2022assembly101} offers a multi-view setting and is larger in scale, our dataset stands out by moving away from lab environments. It includes scenes from uncontrolled settings captured from a third-person view, marking a significant step towards recognizing and analysing  hand-object interactions in the wild.

\subsection{Action Recognition}

Unlike action recognition datasets, which feature pre-segmented clips of high-level activities typically performed by an individual, our dataset contains frame-wise annotations of per-hand manipulation actions performed by multiple people in the scene.

For instance, ActivityNet~\cite{Heilbron2015ActivityNetAL} or Kinetics~\cite{Kay2017TheKH} contains video clips annotated with high-level activities such as painting furniture, washing dishes, driving a car, and planting trees. In these clips, hands are part of the broader context of these actions, rather than the primary focus.
NTU~\cite{ntu_60, ntu_120} includes subsets of hand-centric actions such as "throw", "pick up", and "drop", performed in a scripted manner by individuals in a lab environment. These actions are isolated and lack the complexity of natural interactions.
A subset of actions in AVA-Action~\cite{Gu_2018_CVPR} are hand-related, such as touching, carrying, and holding, annotated at 1Hz.
Hence they lack the start and end of an action. 
Furthermore, the videos are typically low in quality, and contains scenes from movies in rather non-daily life situations,  making it difficult to study hands in detail or spending efforts to provide additional annotations. 

Something-Something~\cite{Goyal2017TheS} contains clips of finer-grained hand actions related to manipulating objects, offering a more focused view on hand-object interactions.
The Charades~\cite{Sigurdsson2016HollywoodIH} dataset contains videos of daily indoor activities where actors were tasked with performing scripted short-term actions, hence lacks the expected natural diversity of hand manipulations found in real situations.

\subsection{Temporal Action Segmentation (TAS)}

Widely used datasets for TAS typically focus on procedural activities, primarily confined to kitchen settings and cooking-related tasks.
GTEA~\cite{Fathi2011LearningTR} is an egocentric dataset featuring videos recorded in a single kitchen, capturing activities such as making sandwiches and preparing coffee.
50Salads~\cite{Stein2013CombiningEA} contains videos of salad preparation captured from a top-down perspective, showcasing activities such as cutting vegetables and mixing ingredients.
Breakfast~\cite{Kuehne2014TheLO} increases the diversity of kitchen environments and individuals performing actions related to breakfast preparation. The scenes are captured from a third-person view using 3 to 5 cameras. The dataset includes finer-level actions such as "taking a cup" and "pouring milk" as part of coarser activities like making coffee.

Our dataset can serve as yet another dataset for TAS, but it goes beyond kitchen environments and activities by focusing on per-hand manipulations in uncontrolled, free-play settings where inter-category and intra-category manipulation segments vary in length, making it a challenging dataset for TAS. 
\section{ChildPlay-Hand}
\label{sec:dataset}
\vspace{-1mm}
In this section, we present ChildPlay-Hand by first defining the hand interaction class labels (Sec.~\ref{sec:interact}), providing the annotation protocol (Sec.~\ref{sec:ann_protocol}), presenting statistics of the annotations (Sec.~\ref{sec:stats}), and comparing it with other datasets (Sec.~\ref{sec:compare_datasets}).

\vspace{-4mm}
\subsection{Hand Interactions }
\label{sec:interact}

We are interested in the main stages of a hand manipulation cycle that a person would consider to grossly description interactions with an object: $background \rightarrow grasp \rightarrow hold / operate \rightarrow release \rightarrow background$; these are complemented with other more precise activities  (see below). 
Additionally, we annotated other actions beyond the aforementioned ones:  the various pre-release gestures differentiating  between different ways of getting rid of an object, such as dropping, putting/placing, throwing, and giving; and   pointing. 
This was done given the nature of our video dataset, which features multi-person scenes, and due to the importance of these gestures for analysing children behaviors and adult-children interaction. In summary, the set of considered categories in ChildPlay-Hand  and their definitions are as follows:

\begin{mdframed}[backgroundcolor=gray!20, hidealllines=true, roundcorner=6pt]
\begin{itemize}
    \item \textbf{\textcolor{black}{Background}:} Hands Idle,  i.e. not holding any object. 
    \item \textbf{\textcolor{red}{Grasp}:} is the transition moment from idle hands to securing an object. It typically starts with a shift in visual attention, followed by approaching and then securing the object.
    \item \textbf{\textcolor{blue}{Hold}:} Passively holding an object in the hand.
    \item \textbf{\textcolor{orange}{Operate}:} Performing an intentional activity with the object (e.g., playing, disassembling, distorting, displacing).
    \item \textbf{\textcolor{violet}{Give}:} Extend your hand towards a person with the intention of giving them an object (whatever the object is taken or not).
    \item \textbf{\textcolor{teal}{Put/Place}:} Place an object on a surface.
    \item \textbf{\textcolor{cyan}{Drop}:} Releasing an object from a distance.
    \item \textbf{\textcolor{olive}{Throw}:} Throw an object.
    \item \textbf{\textcolor{purple}{Release}:} is the transition moment from having the object in hand to having idle hands by simply letting go, dropping, putting/placing, throwing, or giving it to someone (i.e. is situated at the end of the previous actions). 
    \item \textbf{\textcolor{magenta}{Point}:} Pointing at an object, person or area if the object is not countable, e.g., a wall. 
\end{itemize}
\end{mdframed}

The above level of granularity has several advantages. It covers key moments of a complete hand-object interaction cycle without being too fine-grained or coarse. 
For example, grasping could be broken down into micro-events like initiating intention, hand pre-shaping, reach planning, approaching, object contact, and securing the object. 
Similarly, a cycle or series of cycles, depending on the object and activity, can describe actions like assembling a toy, 
but this is not our aim here.  

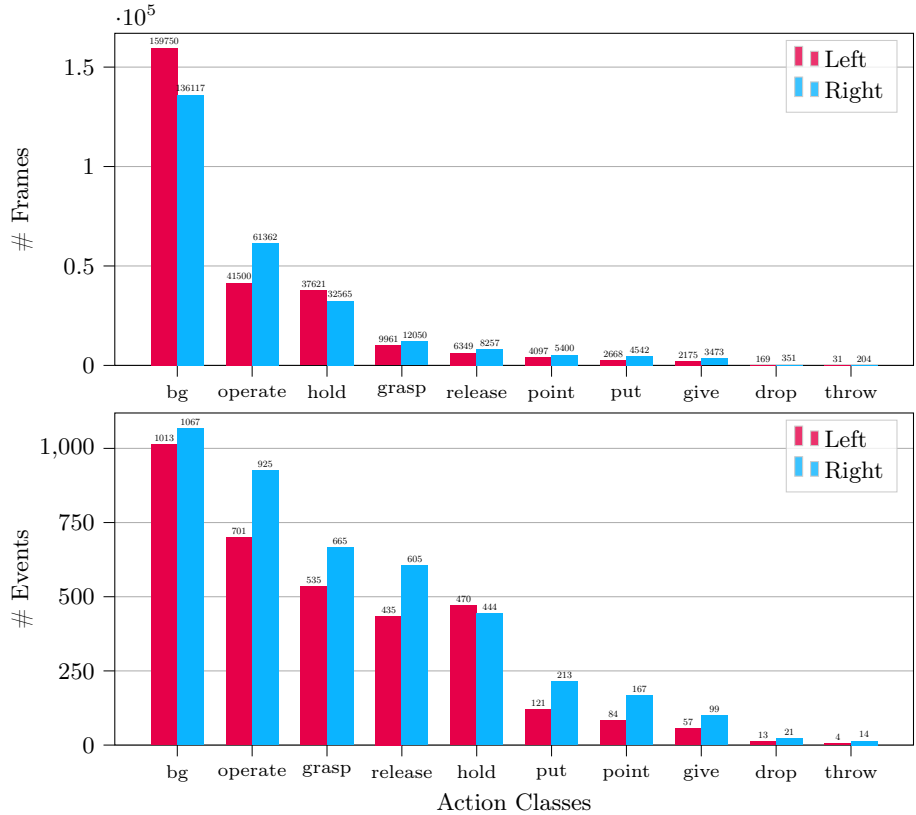
\begin{figure}[t]
\centering

\begin{subfigure}[t]{\textwidth}
\begin{tikzpicture}

\definecolor{crimson230073}{RGB}{230,0,73}
\definecolor{darkgray176}{RGB}{176,176,176}
\definecolor{deepskyblue11180255}{RGB}{11,180,255}
\definecolor{lightgray204}{RGB}{204,204,204}

\begin{axis}[
    width=\textwidth,
    height=6cm,
    legend cell align={left},
    legend style={fill opacity=0.8, draw opacity=1, text opacity=1, draw=lightgray204},
    tick align=outside,
    tick pos=left,
    x grid style={darkgray176},
    xmin=-0.835, xmax=9.835,
    xtick style={color=black},
    xtick={0,1,2,3,4,5,6,7,8,9},
    xticklabels={bg,operate,hold,grasp,release,point,put,give,drop,throw},
    xticklabel style={font=\scriptsize},
    y grid style={darkgray176},
    ylabel={\# Frames},
    ymajorgrids,
    ymin=0, ymax=166996.2,
    ytick style={color=black}
]
\draw[draw=none,fill=crimson230073] (axis cs:-0.35,0) rectangle (axis cs:0,159750);
\addlegendimage{ybar,ybar legend,draw=none,fill=crimson230073}
\addlegendentry{Left}

\draw[draw=none,fill=crimson230073] (axis cs:0.65,0) rectangle (axis cs:1,41500);
\draw[draw=none,fill=crimson230073] (axis cs:1.65,0) rectangle (axis cs:2,37621);
\draw[draw=none,fill=crimson230073] (axis cs:2.65,0) rectangle (axis cs:3,9961);
\draw[draw=none,fill=crimson230073] (axis cs:3.65,0) rectangle (axis cs:4,6349);
\draw[draw=none,fill=crimson230073] (axis cs:4.65,0) rectangle (axis cs:5,4097);
\draw[draw=none,fill=crimson230073] (axis cs:5.65,0) rectangle (axis cs:6,2668);
\draw[draw=none,fill=crimson230073] (axis cs:6.65,0) rectangle (axis cs:7,2175);
\draw[draw=none,fill=crimson230073] (axis cs:7.65,0) rectangle (axis cs:8,169);
\draw[draw=none,fill=crimson230073] (axis cs:8.65,0) rectangle (axis cs:9,31);
\draw[draw=none,fill=deepskyblue11180255] (axis cs:2.77555756156289e-17,0) rectangle (axis cs:0.35,136117);
\addlegendimage{ybar,ybar legend,draw=none,fill=deepskyblue11180255}
\addlegendentry{Right}

\draw[draw=none,fill=deepskyblue11180255] (axis cs:1,0) rectangle (axis cs:1.35,61362);
\draw[draw=none,fill=deepskyblue11180255] (axis cs:2,0) rectangle (axis cs:2.35,32565);
\draw[draw=none,fill=deepskyblue11180255] (axis cs:3,0) rectangle (axis cs:3.35,12050);
\draw[draw=none,fill=deepskyblue11180255] (axis cs:4,0) rectangle (axis cs:4.35,8257);
\draw[draw=none,fill=deepskyblue11180255] (axis cs:5,0) rectangle (axis cs:5.35,5400);
\draw[draw=none,fill=deepskyblue11180255] (axis cs:6,0) rectangle (axis cs:6.35,4542);
\draw[draw=none,fill=deepskyblue11180255] (axis cs:7,0) rectangle (axis cs:7.35,3473);
\draw[draw=none,fill=deepskyblue11180255] (axis cs:8,0) rectangle (axis cs:8.35,351);
\draw[draw=none,fill=deepskyblue11180255] (axis cs:9,0) rectangle (axis cs:9.35,204);
\draw (axis cs:-0.175,159750) node[
  scale=0.4,
  anchor=south,
  text=black,
  rotate=0.0
]{159750};
\draw (axis cs:0.825,41500) node[
  scale=0.4,
  anchor=south,
  text=black,
  rotate=0.0
]{41500};
\draw (axis cs:1.825,37621) node[
  scale=0.4,
  anchor=south,
  text=black,
  rotate=0.0
]{37621};
\draw (axis cs:2.825,9961) node[
  scale=0.4,
  anchor=south,
  text=black,
  rotate=0.0
]{9961};
\draw (axis cs:3.825,6349) node[
  scale=0.4,
  anchor=south,
  text=black,
  rotate=0.0
]{6349};
\draw (axis cs:4.825,4097) node[
  scale=0.4,
  anchor=south,
  text=black,
  rotate=0.0
]{4097};
\draw (axis cs:5.825,2668) node[
  scale=0.4,
  anchor=south,
  text=black,
  rotate=0.0
]{2668};
\draw (axis cs:6.825,2175) node[
  scale=0.4,
  anchor=south,
  text=black,
  rotate=0.0
]{2175};
\draw (axis cs:7.825,169) node[
  scale=0.4,
  anchor=south,
  text=black,
  rotate=0.0
]{169};
\draw (axis cs:8.825,31) node[
  scale=0.4,
  anchor=south,
  text=black,
  rotate=0.0
]{31};
\draw (axis cs:0.175,136117) node[
  scale=0.4,
  anchor=south,
  text=black,
  rotate=0.0
]{136117};
\draw (axis cs:1.175,61362) node[
  scale=0.4,
  anchor=south,
  text=black,
  rotate=0.0
]{61362};
\draw (axis cs:2.175,32565) node[
  scale=0.4,
  anchor=south,
  text=black,
  rotate=0.0
]{32565};
\draw (axis cs:3.175,12050) node[
  scale=0.4,
  anchor=south,
  text=black,
  rotate=0.0
]{12050};
\draw (axis cs:4.175,8257) node[
  scale=0.4,
  anchor=south,
  text=black,
  rotate=0.0
]{8257};
\draw (axis cs:5.175,5400) node[
  scale=0.4,
  anchor=south,
  text=black,
  rotate=0.0
]{5400};
\draw (axis cs:6.175,4542) node[
  scale=0.4,
  anchor=south,
  text=black,
  rotate=0.0
]{4542};
\draw (axis cs:7.175,3473) node[
  scale=0.4,
  anchor=south,
  text=black,
  rotate=0.0
]{3473};
\draw (axis cs:8.175,351) node[
  scale=0.4,
  anchor=south,
  text=black,
  rotate=0.0
]{351};
\draw (axis cs:9.175,204) node[
  scale=0.4,
  anchor=south,
  text=black,
  rotate=0.0
]{204};
\end{axis}

\end{tikzpicture}
\end{subfigure}

\begin{subfigure}[t]{\textwidth}
\begin{tikzpicture}

\definecolor{crimson230073}{RGB}{230,0,73}
\definecolor{darkgray176}{RGB}{176,176,176}
\definecolor{deepskyblue11180255}{RGB}{11,180,255}
\definecolor{lightgray204}{RGB}{204,204,204}

\begin{axis}[
    width=\textwidth,
    height=6cm,
    legend cell align={left},
    legend style={fill opacity=0.8, draw opacity=1, text opacity=1, draw=lightgray204},
    tick align=outside,
    tick pos=left,
    x grid style={darkgray176},
    xlabel={Action Classes},
    xmin=-0.835, xmax=9.835,
    xtick style={color=black},
    xtick={0,1,2,3,4,5,6,7,8,9},
    xticklabels={bg,operate,grasp,release,hold,put,point,give,drop,throw},
    xticklabel style={font=\scriptsize},
    y grid style={darkgray176},
    ylabel={\# Events},
    ymajorgrids,
    ymin=0, ymax=1119.3,
    ytick={0,250,500,750,1000},
    ytick style={color=black}
]

\draw[draw=none,fill=crimson230073] (axis cs:-0.35,0) rectangle (axis cs:0,1013);
\addlegendimage{ybar,ybar legend,draw=none,fill=crimson230073}
\addlegendentry{Left}

\draw[draw=none,fill=crimson230073] (axis cs:0.65,0) rectangle (axis cs:1,701);
\draw[draw=none,fill=crimson230073] (axis cs:1.65,0) rectangle (axis cs:2,535);
\draw[draw=none,fill=crimson230073] (axis cs:2.65,0) rectangle (axis cs:3,435);
\draw[draw=none,fill=crimson230073] (axis cs:3.65,0) rectangle (axis cs:4,470);
\draw[draw=none,fill=crimson230073] (axis cs:4.65,0) rectangle (axis cs:5,121);
\draw[draw=none,fill=crimson230073] (axis cs:5.65,0) rectangle (axis cs:6,84);
\draw[draw=none,fill=crimson230073] (axis cs:6.65,0) rectangle (axis cs:7,57);
\draw[draw=none,fill=crimson230073] (axis cs:7.65,0) rectangle (axis cs:8,13);
\draw[draw=none,fill=crimson230073] (axis cs:8.65,0) rectangle (axis cs:9,4);
\draw[draw=none,fill=deepskyblue11180255] (axis cs:2.77555756156289e-17,0) rectangle (axis cs:0.35,1067);
\addlegendimage{ybar,ybar legend,draw=none,fill=deepskyblue11180255}
\addlegendentry{Right}

\draw[draw=none,fill=deepskyblue11180255] (axis cs:1,0) rectangle (axis cs:1.35,925);
\draw[draw=none,fill=deepskyblue11180255] (axis cs:2,0) rectangle (axis cs:2.35,665);
\draw[draw=none,fill=deepskyblue11180255] (axis cs:3,0) rectangle (axis cs:3.35,605);
\draw[draw=none,fill=deepskyblue11180255] (axis cs:4,0) rectangle (axis cs:4.35,444);
\draw[draw=none,fill=deepskyblue11180255] (axis cs:5,0) rectangle (axis cs:5.35,213);
\draw[draw=none,fill=deepskyblue11180255] (axis cs:6,0) rectangle (axis cs:6.35,167);
\draw[draw=none,fill=deepskyblue11180255] (axis cs:7,0) rectangle (axis cs:7.35,99);
\draw[draw=none,fill=deepskyblue11180255] (axis cs:8,0) rectangle (axis cs:8.35,21);
\draw[draw=none,fill=deepskyblue11180255] (axis cs:9,0) rectangle (axis cs:9.35,14);
\draw (axis cs:-0.175,1013) node[
  scale=0.4,
  anchor=south,
  text=black,
  rotate=0.0
]{1013};
\draw (axis cs:0.825,701) node[
  scale=0.4,
  anchor=south,
  text=black,
  rotate=0.0
]{701};
\draw (axis cs:1.825,535) node[
  scale=0.4,
  anchor=south,
  text=black,
  rotate=0.0
]{535};
\draw (axis cs:2.825,435) node[
  scale=0.4,
  anchor=south,
  text=black,
  rotate=0.0
]{435};
\draw (axis cs:3.825,470) node[
  scale=0.4,
  anchor=south,
  text=black,
  rotate=0.0
]{470};
\draw (axis cs:4.825,121) node[
  scale=0.4,
  anchor=south,
  text=black,
  rotate=0.0
]{121};
\draw (axis cs:5.825,84) node[
  scale=0.4,
  anchor=south,
  text=black,
  rotate=0.0
]{84};
\draw (axis cs:6.825,57) node[
  scale=0.4,
  anchor=south,
  text=black,
  rotate=0.0
]{57};
\draw (axis cs:7.825,13) node[
  scale=0.4,
  anchor=south,
  text=black,
  rotate=0.0
]{13};
\draw (axis cs:8.825,4) node[
  scale=0.4,
  anchor=south,
  text=black,
  rotate=0.0
]{4};
\draw (axis cs:0.175,1067) node[
  scale=0.4,
  anchor=south,
  text=black,
  rotate=0.0
]{1067};
\draw (axis cs:1.175,925) node[
  scale=0.4,
  anchor=south,
  text=black,
  rotate=0.0
]{925};
\draw (axis cs:2.175,665) node[
  scale=0.4,
  anchor=south,
  text=black,
  rotate=0.0
]{665};
\draw (axis cs:3.175,605) node[
  scale=0.4,
  anchor=south,
  text=black,
  rotate=0.0
]{605};
\draw (axis cs:4.175,444) node[
  scale=0.4,
  anchor=south,
  text=black,
  rotate=0.0
]{444};
\draw (axis cs:5.175,213) node[
  scale=0.4,
  anchor=south,
  text=black,
  rotate=0.0
]{213};
\draw (axis cs:6.175,167) node[
  scale=0.4,
  anchor=south,
  text=black,
  rotate=0.0
]{167};
\draw (axis cs:7.175,99) node[
  scale=0.4,
  anchor=south,
  text=black,
  rotate=0.0
]{99};
\draw (axis cs:8.175,21) node[
  scale=0.4,
  anchor=south,
  text=black,
  rotate=0.0
]{21};
\draw (axis cs:9.175,14) node[
  scale=0.4,
  anchor=south,
  text=black,
  rotate=0.0
]{14};
\end{axis}

\end{tikzpicture} 
\end{subfigure}

\caption{Distribution of hand action classes in the dataset. We show the distribution in frames (top) and events (bottom).}
\label{fig:childplay-hand-stats} \vspace{-0.5cm}
\end{figure}

\begin{figure}[t]
\centering
\begin{subfigure}[b]{\textwidth}
   \input{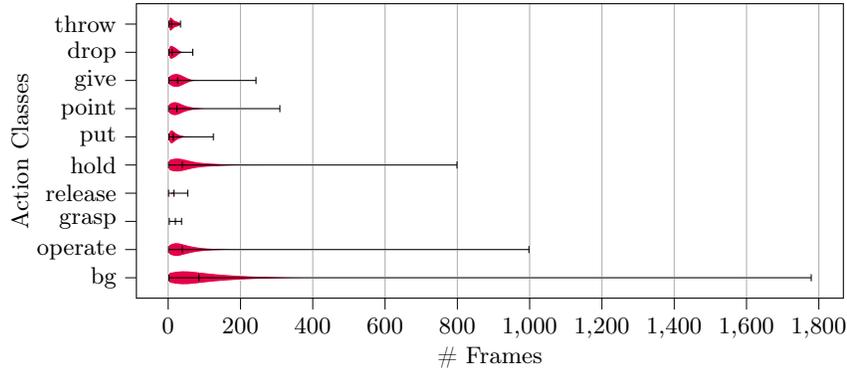}
\end{subfigure} \vspace{-0.7cm}
\caption{Distribution of event duration (in frames) per action class. The violin plot shows the min, max and median values of each distribution.}
\label{fig:childplay-hand-duration-stats} \vspace{-0.6cm}
\end{figure} \vspace{-0.4cm}

\subsection{Annotation Protocol}
\label{sec:ann_protocol}
\vspace{-1mm}

\mypartitle{Source of Videos: }We borrowed video clips from ChildPlay-Gaze~\cite{tafasca2023childplay} and annotated them with the defined hand actions described in Sec.~\ref{sec:interact}. 
This dataset was originally annotated with the 2D pixel location of gaze targets for the task of gaze following, focusing on children's gaze. 
The raw videos were downloaded from YouTube and feature children playing and interacting with adults in uncontrolled environments such as childcare facilities, schools, homes, and therapy centers. The dataset contains 401 video clips of high visual quality, mainly in indoor settings, featuring at least one child, often with one or two adults and multiple children. Activities are unscripted, with the dominant activity typically being "playing with toys".

\mypartitle{Annotation Process:}
We densely annotated ChildPlay-Gaze clips with body bounding boxes, per-hand interactions with objects, involved object bounding boxes, for up to three people in the scene, the same as those annotated with gaze. 
Seven annotators were assigned on the annotation process, followed by a review stage done by an expert in the field. 
\sam{The two action classes of \texttt{grasp} and \texttt{release} were not manually annotated because they can be inferred from the existing annotations at the transition moments (\eg a person's hand changing from \texttt{background} to \texttt{hold} or \texttt{operate} has inevitably gone through a \texttt{grasp} event). 
We empirically set the duration for \texttt{grasp} to 20 frames, or approximately $0.6s$, with 4 frames inside the object manipulation segment and 16 frames outside (\ie \texttt{background}). For \texttt{release}, this duration is split into 8 frames inside and 8 frames outside. When the duration of the preceding (resp. following) event for \texttt{grasp} (resp. \texttt{release}) is smaller than the outside frames (16 for \texttt{grasp} and 8 for \texttt{release}), we use the available frames instead. We conduct a manual verification step afterwards to ensure that the \texttt{grasp} and \texttt{release} instances are valid and acceptable.
}

\vspace{-1mm}
\subsection{Annotation Statistics}
\label{sec:stats}
\vspace{-1mm}

Figure \ref{fig:childplay-hand-stats} provides the per-hand distribution of action labels by frames and by event segments. Expectedly, the distribution features a long tail with actions like \texttt{give}, \texttt{drop}, and \texttt{throw} having much fewer instances than the rest. 
It is also interesting to note that the right hand \texttt{operate} has a higher frequency than the left counterpart, whereas the opposite can be observed for the \texttt{hold} class. 
This is likely due to the fact that a higher percentage of people is right-handed, so the active manipulation of objects is more often associated with the dominant right hand first, while the more passive \texttt{hold} is delegated to the left. 
We follow the training, validation, and test splits in ChildPlay-Gaze~\cite{tafasca2023childplay}. More information about the dataset splits can be found in the supplementary material.

\sam{We also provide the frame-wise  event duration distribution of each action class  (\cf Figure \ref{fig:childplay-hand-duration-stats}). We observe that, aside from the \texttt{background} class which tends to be longer, all other actions typically last between $0$ and $4$ seconds (assuming a frame rate of $30$ FPS), with holding and operating being relatively longer compared to the other more short-term and specific labels.}

\vspace{-2mm}

\subsection{Comparison to Other Datasets}
\label{sec:compare_datasets}

\vspace{-1mm}

Tab.~\ref{tab:dataset} compares ChildPlay-Hand across different aspects. While our dataset has fewer segments and actions due to the granularity we focus on, it has other unique aspects. 
ChildPlay-Hand features per-hand actions, captures in-the-wild scenes of children playing in a free-play manner in various backgrounds, and includes multiple people interacting with the same objects.

\begin{table}[t]
\centering
\resizebox{\linewidth}{!}{
\begin{tabular}{l|c|c|c|c|c|c}
\hline
\textbf{Dataset} & \textbf{View} & \textbf{\# Segments} & \textbf{\# Actions} & \textbf{Per-Hands} & \textbf{In-the-Wild} & \textbf{Multi-Person}  \\
\hline
50Salads~\cite{Stein2013CombiningEA} & Top-Down & 899 & 6  & \xmark & \xmark & \xmark \\
Breakfast~\cite{Kuehne2014TheLO}     & Exo & 11,300 & 14 & \xmark & \xmark & \xmark \\
Assembly101-Coarse~\cite{sener2022assembly101} & Ego-Exo & 104,759 & 11 & \xmark & \xmark & \xmark \\
Assembly101-Fine~\cite{sener2022assembly101} & Ego-Exo & $~$1M & 24 & \xmark & \xmark & \xmark \\
\textbf{ChildPlay-Hand}  & Exo & 7,653 & 10 & \cmark & \cmark & \cmark \\
\hline
\end{tabular}
}
\caption{
Comparison of ChildPlay-Hand with other related datasets. Note that \# Actions refers to verbs in other datasets.}
\label{tab:dataset} \vspace{-0.7cm}
\end{table}
\vspace{-0.4cm}

\section{Experiments}
\label{sec:experiments}

\vspace{-3mm}

In this section, we define the tasks (Sec.~\ref{sec:tasks}), discuss the selected models, protocol, implementation details for recognition (Sec.~\ref{sec:action_rec}) and segmentation (Sec.~\ref{sec:tas_methods}). 

\vspace{-4mm}

\subsection{Benchmarked Tasks }
\label{sec:tasks}

\vspace{-1mm}

The ChildPlay-Hand dataset can be used for various video understanding tasks. Following the standard terminology, we can address Action Recognition by exploiting pre-segmented actions. It can also be used for Spatio-Temporal Action Localization, similar to the AVA-Action dataset~\cite{Gu_2018_CVPR}, which involves not only recognizing but also localizing actions performed by all individuals within a key frame. 
Alternatively, the task can be formulated as an Human-Object Interaction recognition problem, by predicting all triplets within a key frame, <person bounding box, hand interaction, object bounding box>.

In this work, we focus on \textbf{Temporal Action Segmentation (TAS)}, where the goal is to perform frame-wise prediction  \( Y = \{y_1, \ldots, y_T\} \) on an input video \( X = \{x_1, \ldots, x_T\} \) of a person where, \(y_t \in \{\text{$a_{L}$}, \text{$a_{R}$}\}\)
denote the left and right hand actions 
and  can belong to one of \( C \) categories.

\mypartitle{T1: Object in Hand (OiH):} 
We first investigate the binary task of detecting  whether a person has an object in a given hand. 
We generate labels by categorizing \texttt{hold} and \texttt{operate} as positive (OiH) and the rest as negative. 
The main motivation behind this task is that it serves as a prequel to segmenting different stages between non-OiH and OiH and  evaluates the ability of a model to detect the presence of an object in hand. This task remains very challenging due to self-occlusions or hands being close (i.e., objects being occluded by hands), the small sizes and great variety  of objects.

\mypartitle{T2: Manipulation Stages (ManiS):} Next, we target the primary stages of an object manipulation cycle: hands begin in an idle state (\texttt{background}), grasp an object (\texttt{grasp}), perform a series of holding (\texttt{hold}) and operating actions (\texttt{operate}), release the object (\texttt{release}), and return to the idle state. To distinguish between these categories and segment them, a combination of two primary cues is needed: distinguishing among the hand motions and tracking the presence or absence of an object in the hands, thus making it challenging. 

To perform frame-wise prediction of the entire input sequence, methods in TAS usually rely on pre-extracted features for each frame, e.g. using  I3D~\cite{Carreira2017QuoVA}. However, these representations need to be informative and tailored to the task. 
To this end, we explore different spatio-temporal networks (detailed in Sec.~\ref{sec:action_rec}) and fine-tune them on ChildPlay-Hand under a recognition protocol to measure their ability to recognize object and hand manipulations within a short window. Then, we use the top-performing recognition network to extract frame-wise representations for the TAS methods in Sec.~\ref{sec:tas_methods}. 

\vspace{-3mm}
\subsection{OiH and ManiS Recognition}
\label{sec:action_rec}
\vspace{-1mm}

Below, we first detail the two types of video volume used as input to the networks and then provide a brief overview of the selected networks. 

\mypartitle{Inputs:} 
We experiment with two types of inputs for the person of interest (PoI) in the scene: full-body and hand-region. 
For each type, we use the corresponding per-frame bounding boxes to crop out the body or hand of the PoI according to the Subject-Centered Cropping method from~\cite{Duan_2022_CVPR}:  we compute the body/hand bounding box that encapsulates the PoI across frames in a short temporal window and use it to crop the input modalities, such as RGB frames and keypoints, across all frames in the window. 
For the body variant, we use the available ground-truth bounding boxes.
For the hand, as ground-truth boxes are unavailable, we generate pseudo-hand bounding boxes as described in the supp. materials with illustrations.

\mypartitle{Methods:}  PoseConv3D~\cite{Duan_2022_CVPR} is a strong state-of-the-art (SoA) model for skeleton-based action recognition. We experiment with PoseConv3D as a baseline to evaluate the performance of pose-only representation. Unlike graph-based methods~\cite{Li2018SpatioTemporalGC} that use the coordinate-based representation of 3D body joints, PoseConv3D creates heatmap volumes of 2D body joints and has shown to be more robust. Each heatmap has a size of $K \times H \times W$. Here, $K$ denotes the number of joints, represented by Gaussian maps centered at each joint. These heatmaps are then stacked along the temporal dimension to form a volume of size $K \times T \times H \times W$. Then, the input heatmap volumes are encoded using 3D-CNNs. For body keypoints, we extract \(K=17\) keypoints using HRNet~\cite{Wang2019DeepHR}, and for hand keypoints, we use \(K=21\) per hand, extracted using ZoomNet~\cite{jin2020whole}. Note, to make use of the pre-trained PoseConv3D weights, which are trained on body keypoints, for use with hand keypoint heatmaps, we add an adapter layer to the model that maps the 21 channels of hand keypoints to the 17 channels expected by PoseConv3D, using a 3D-CNN layer.

However, pose alone (esp. body pose) is insufficient to recognize actions such as grasping, holding, and releasing, as these actions require appearance cues to determine whether objects are in the hands. 
Therefore, we experiment with the multimodal version of PoseConv3D, RGBPoseConv3D~\cite{Duan_2022_CVPR}, which has two pathways: a pose pathway operating at a higher frame rate and a visual pathway using RGB frames at a lower frame rate, similar to Slow-fast networks~\cite{Feichtenhofer2018SlowFastNF}. The fast branch for pose captures fine-grained motion of the keypoints, while the slow visual branch can provide appearance context (e.g., objects in the hands) with lateral connections between the two pathways. Following~\cite{Duan_2022_CVPR}, we use 8 frames with a temporal stride of 4 for the visual pathway and 32 frames with no stride for the pose pathway.

Additionally, given the success of transformer-based architectures in video understanding, we experiment with a recent hierarchical transformer, Hiera~\cite{ryali2023hiera}. This network has demonstrated strong performance, particularly on the AVA-Action~\cite{Gu_2018_CVPR} dataset, as also shown in LART~\cite{Rajasegaran2023OnTB}. Please refer to the supp. materials for the implementation details of these networks.

\mypartitle{Recognition Metrics: }For the recognition part, we compare models based on frame-based metrics only. For the OiH task, we use accuracy as well as precision, recall, and F1 of the OiH state. For the ManiS task, we compute accuracy as well as macro precision, recall, and F1 by averaging each of these metrics across the categories. The choice of macro metrics is due to the imbalanced nature of the dataset, treating all classes equally since they all represent main stages of hand-object manipulation.

\vspace{-2mm}
\subsection{OiH and ManiS Segmentation}
\label{sec:tas_methods}
\vspace{-2mm}

\mypartitle{Input: }TAS methods process entire sequences as input and make frame-wise predictions. They rely on extracted features from spatio-temporal networks trained for recognition to form a feature grid sized $L \times D$, where $L$ represents the sequence length and $D$ denotes the feature dimension.

\mypartitle{Methods:}
As a first baseline, we use the best frame-recognition network identified in Sec.~\ref{sec:action_rec} to produce frame-wise prediction via a standard Sliding-Window strategy. 
However, to account for a larger temporal context, smooth predictions and avoid over-segmentation, and learn the plausibility of manipulation activity sequence and ordering, we experiment with MS-TCN~\cite{AbuFarha2019MSTCNMT}, a multi-stage convolutional-based network specifically designed for TAS. 
It relies on 1D convolutional kernels to process the feature grid along time, maintaining full temporal resolution across its multi-stage architecture. 
Each stage (SS-TCN) of MS-TCN  comprises multiple layers of dilated convolutions, which are used to progressively increase the receptive field
and to refine the predictions made by the previous stage.

\mypartitle{Segmentation Metrics:}
In addition to the frame-based metrics, we compute segmental metrics. 
Given the predicted and ground-truth segments, we first find the optimal matching between them using Bipartite matching~\cite{Kuhn1955TheHM}. 
The cost function used is as follows:
\[
C(i, j) =
\begin{cases}
1 - O(D_i, G_j) & \text{if } label(D_i) = label(G_j) \\
2 & \text{if } label(D_i) \neq label(G_j)
\end{cases}
\]
where \( O(D_i, G_j) \) is the overlap between a predicted segment \( D_i \) and a ground-truth segment \( G_j \), defined as:
\[
O(D_i, G_j) = 2 \times \frac{\text{Prec.} \times \text{Rec.}}{\text{Prec.} + \text{Rec.}},
\quad
\text{Prec.} = \frac{|D_{i} \cap G_{j}|}{|D_{i}|}, \quad \text{Rec.} = \frac{|D_{i} \cap G_{j}|}{|G_{j}|}
\]

After optimal matching, a matched segment is said to be a true positive if its overlap is above 0 (and of the same class); otherwise, it is considered a false positive. 
Unmatched recognized segments become false positives, and unmatched ground-truth segments become false negatives. Using these counts, we compute precision, recall, and F1. We also report Edit score as in TAS methods~\cite{Ding2022TemporalAS}. 

\vspace{-2mm}
\vspace{-1mm}

\section{Results}
\label{sec:results}
\vspace{-2mm}

In this section, we first present the results of recognition experiments in Sec.~\ref{sec:rec_results}, followed by segmentation results in Sec.~\ref{sec:seg_results}.

\vspace{-1mm}

\begin{table}[t]
\centering
\resizebox{\linewidth}{!}{
\begin{tabular}{lc*{12}{c}}
\hline
Method & Input & N.~Param. & Pre-train & Window & \multicolumn{4}{c}{\textbf{T1: OiH}} & & \multicolumn{4}{c}{\textbf{T2: ManiS}} \\ 
\cline{6-9} \cline{11-14}
 & & & & & Acc. & F1 & Prec. & Rec. & & Acc. & m-F1 & m-Prec. & m-Rec. \\ 
\hline
\textcolor{gray}{Dummy Classifier} & - & - & - & - &  53.6 &   35.3 &  31.4 & 40.3 &  & 40.0 &  19.4 &  19.8 &  19.7 \\
\textcolor{red}{PoseConv3D}~\cite{Duan_2022_CVPR} & body & 2M & NTU-60 & $48\times1$ &  66.5 &  32.6 &  44.3 &  25.8 &  & 63.0 &  30.1 &  36.0 & 28.9  \\
\textcolor{red}{PoseConv3D}~\cite{Duan_2022_CVPR} & hand & 2M & NTU-60 & $48\times1$ & 62.6  &  51.5	 &  43.5 &  63.2 &  & 59.3	 &  32.9 & 34.8 &  31.7 \\
\textcolor{blue}{RGBPoseConv3D}~\cite{Duan_2022_CVPR} & body & 36M & NTU-60 & $8\times4$, $32\times1$ & 72.4 & 51.5	 &  57.7 &  46.6 &  & 66.0 &   33.8 & 44.9	 &  31.2 \\
\textcolor{blue}{RGBPoseConv3D}~\cite{Duan_2022_CVPR} & hand & 36M & NTU-60 & $8\times4$, $32\times1$ & 75.1	 & 53.0 &  65.2 &  44.7 &  & 67.7 &   36.8 &  \textbf{51.1} &   33.4\\
\textcolor{teal}{Hiera}~\cite{ryali2023hiera} & body & 52M & K400 & $16\times2$ & 79.5 &  65.4 & 69.8 &  61.5 &  & 66.5 & 45.0  & 45.2 & 45.7  \\
\textcolor{teal}{Hiera}~\cite{ryali2023hiera} & hand & 52M & K400 & $16\times2$  &  \textbf{85.4} &  \textbf{74.3} & \textbf{83.4} &  \textbf{67.0} &  & \textbf{73.2} &  \textbf{51.2} &  50.6 &  \textbf{53.1} \\
\hline
\end{tabular}
}
\caption{Comparison of different networks. All metrics are frame-based.  m-\{metric\} stands for macro. Stratified sampling is used as dummy classifier. Window of e.g., $16\times2$ refers to 16 frames with a temporal stride of 2.}
\label{tab:main_table} \vspace{-0.4cm}
\end{table}

\begin{figure}[t]
    \centering
    \begin{minipage}{0.32\textwidth}
        \includegraphics[width=\linewidth]{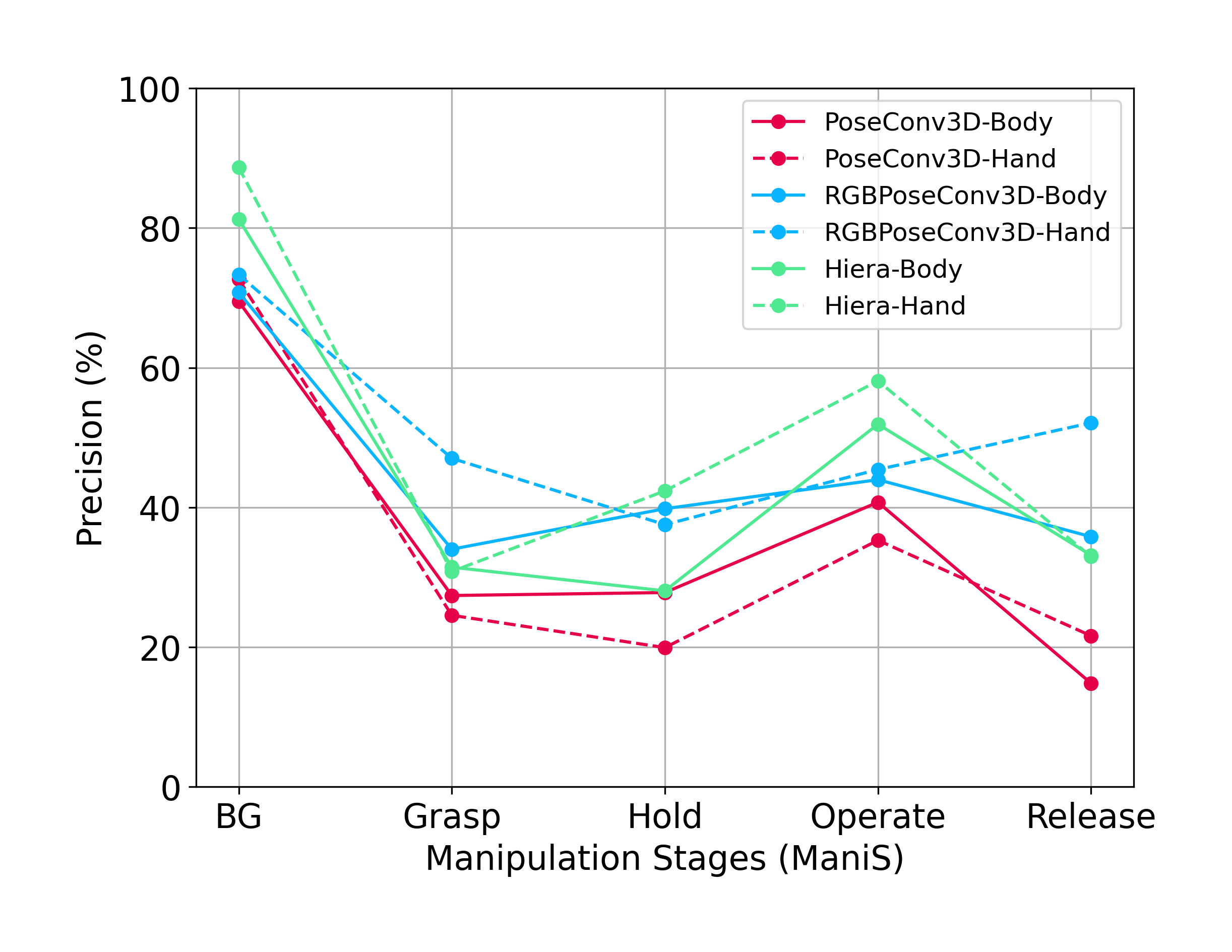}
    \end{minipage}
    \hfill
    \begin{minipage}{0.32\textwidth}
        \includegraphics[width=\linewidth]{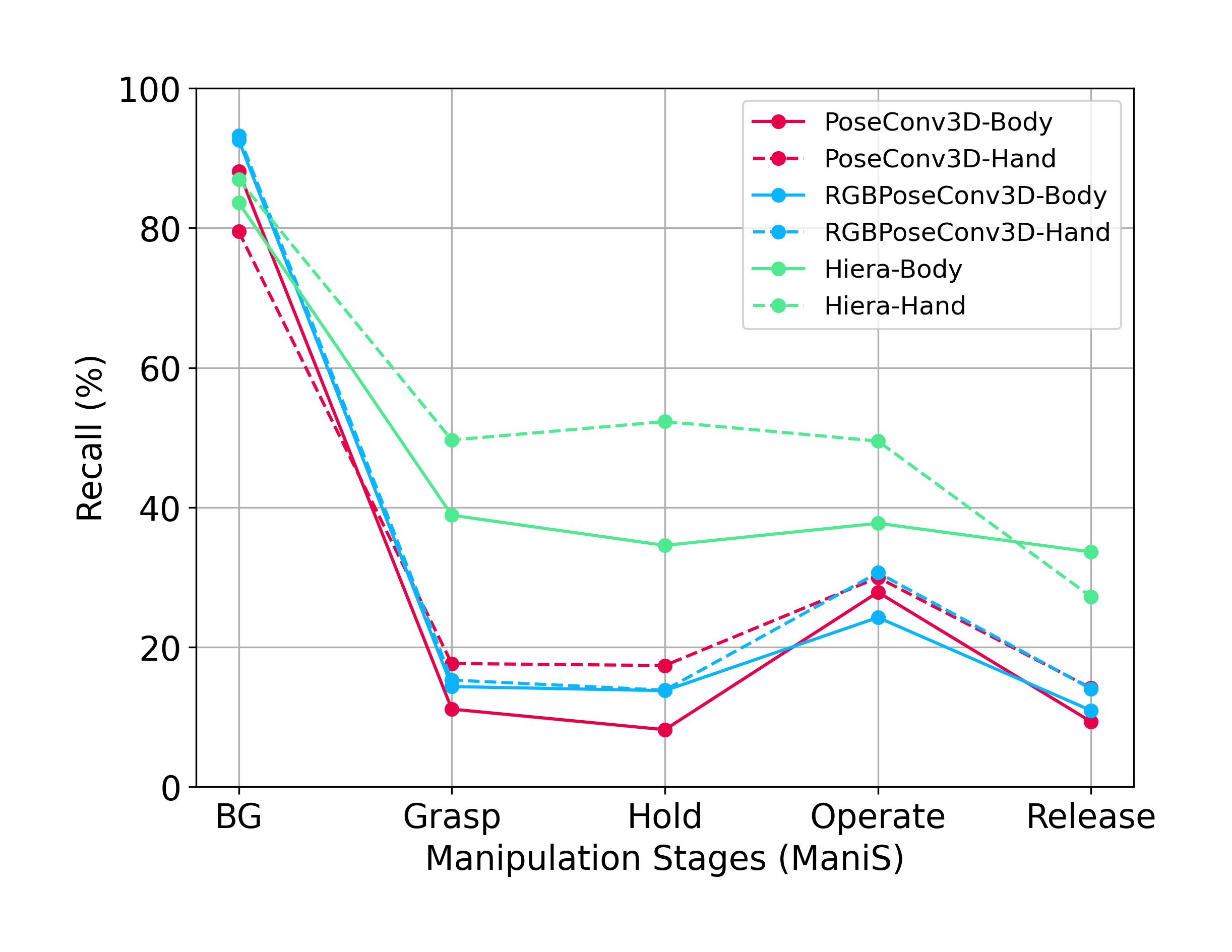}
    \end{minipage}
    \hfill
    \begin{minipage}{0.32\textwidth}
        \includegraphics[width=\linewidth]{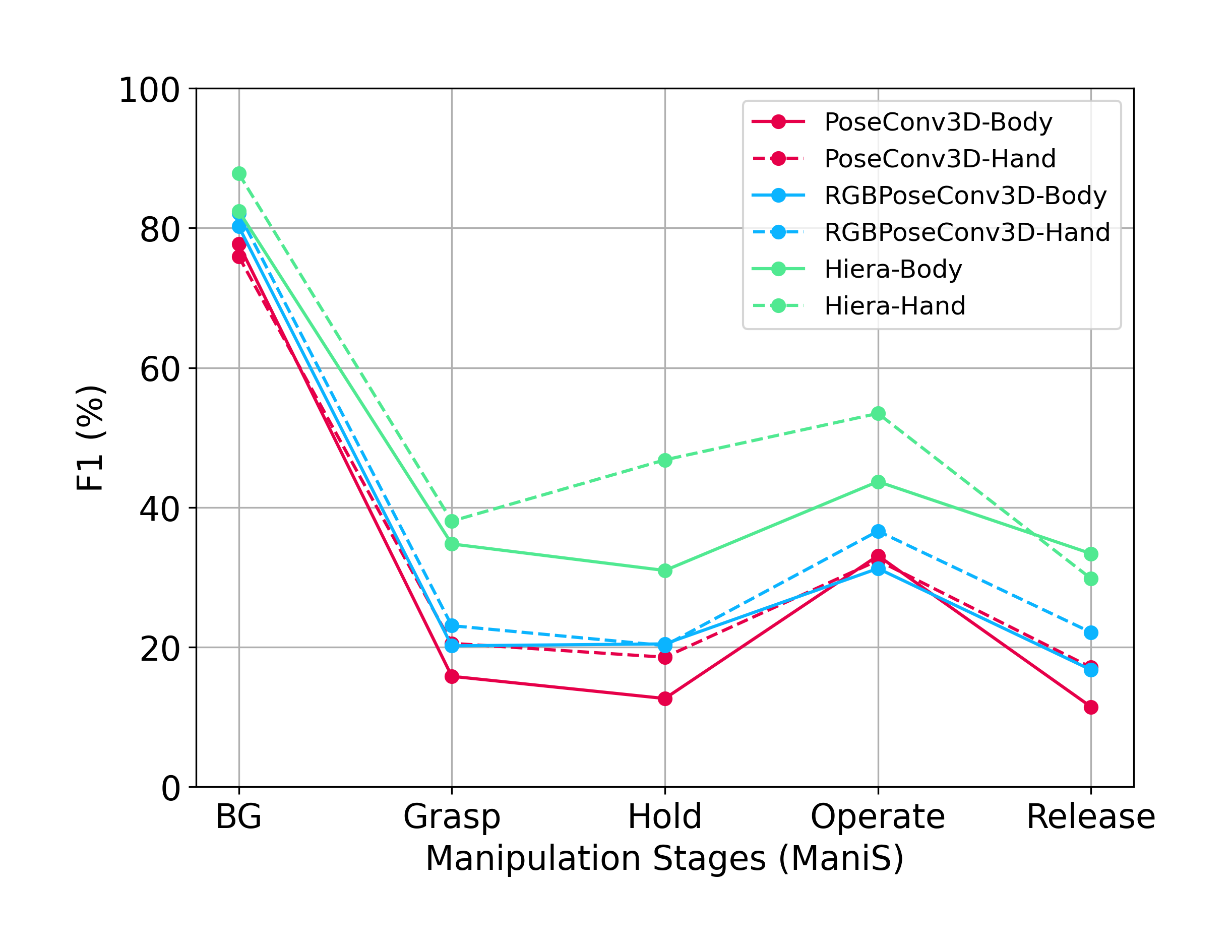}
    \end{minipage}
\vspace{-3mm}
    \caption{Class-wise frame-based: Precision, Recall, and F1.}
    \label{fig:metrics_comparison_frame} 
    \vspace{-0.4cm}
\end{figure}
\vspace{-0.4cm}

\subsection{Recognition Results}
\label{sec:rec_results}
\vspace{-2mm}

\mypartitle{OiH Task:}
As shown in Tab.~\ref{tab:main_table}, the top-performing network  is Hiera with hand inputs, achieving an accuracy of 85.4\% and an F1 score of 74.3\%, 
and  outperforming other methods by a large margin in terms of F1. 
Beyond architectural differences, this performance may also be attributed to the pre-training on 
Kinetics-400 (K400)~\cite{Kay2017TheKH} which offers more diverse scenes than NTU-60~\cite{ntu_60}.

\mypartitle{ManiS Task:}
Hiera applied on hand crops also outperforms the other approaches in this task, with the hand version leading at 73.2\% accuracy and a 51.2\% macro F1 score. However, RGBConvPose3D with hand inputs shows slightly better m-Prec. (+0.5\%). Looking at class-wise performance in Fig.~\ref{fig:metrics_comparison_frame}, the Hiera family generally delivers the best results across manipulation stages, particularly in the \texttt{hold} and \texttt{operate} stages, where the hand version significantly outperforms others. However, other methods like RGBPoseConv3D-Hand exhibit better precision in the \texttt{grasp} and \texttt{release} stages.  Despite Hiera-Hand achieving relatively better performances in the \texttt{hold} and \texttt{operate} stages with F1 scores above 40\%, the transitional stages of \texttt{grasp} and \texttt{release} remain challenging. Confusion matrices can be found in the supp. materials.

\mypartitle{Comparing full-body and hand-region inputs:}
In both tasks, hand variants consistently outperform their body counterparts in terms of F1 score. Looking at Fig.~\ref{fig:metrics_comparison_frame}, this difference is more pronounced in certain classes depending on the method. For instance, Hiera-Hand outperforms its body variant more notably in the \texttt{hold} and \texttt{operate} stages, while the body version performs better on \texttt{release} with higher recall. In the case of RGBPoseConv3D, the hand variant shows notable improvements in the \texttt{operate} and \texttt{release} stages, while performing on par with the body variant on \texttt{hold}. 

Indeed, hand-regions allow models to focus more directly on the main area of interest, which likely contributes to their improved performance. 
However, the body input also contains additional contextual information that could be valuable. 
For instance, the orientation and posture of the body, along with attentional cues, could help distinguish between intentional and unintentional hand movements, which is crucial for differentiating between \texttt{hold} and \texttt{operate}.
It would be interesting to explore ways to leverage both hand-focused and full-body information together. We leave this for future work.

\mypartitle{On the effectiveness of pose-only recognition:}
In OiH task, using pose-alone especially body-pose is indeed insufficient to recognize OiH state. This is reflected in Tab.~\ref{tab:main_table}, where RGBPoseConv3D-Body outperforms PoseConv3D-Body by a large margin in terms of F1 score. 
The same applies to RGBPoseConv3D-Hand and PoseConv3D-Hand, though with a smaller difference, likely due to the fact that hand pose implicitly contains information about objects in hand to some extent, whereas body-pose (with only the wrist keypoint as hand information) does not.
Interestingly, PoseConv3D-Hand performs on par with RGBPoseConv3D-Body, indicating the effectiveness of hand-pose keypoints. In 
ManiS task, we can see a similar trend with RGBPoseConv3D outperforming PoseConv3D. 

Looking at F1 scores in Fig.~\ref{fig:metrics_comparison_frame}, with the body inputs, having appearance cues helps more notably with \texttt{release}, \texttt{hold}, and \texttt{grasp} while pose-only performs slightly better on \texttt{operate}. This shows that while body-pose can be effective when manipulating an object, other stages indeed need appearance cues. With hand inputs, appearance cues help more with \texttt{operate} and \texttt{release} while performing more closely to each other on \texttt{hold} and \texttt{grasp}.

\begin{table}[t]
\centering
\resizebox{\linewidth}{!}{
\begin{tabular}{l*{14}{c}}
\hline
Method &  \multicolumn{6}{c}{\textbf{T1: OiH}} & & \multicolumn{6}{c}{\textbf{T2: ManiS}} \\ 
\cline{2-7} \cline{9-14}
 & S-Prec. & S-Rec. & S-F1 & F-Acc. & F-Prec. & F-Rec. & & m-S-Prec. & m-S-Rec. & m-S-F1 & Edit & F-Acc. & m-F-F1 \\ 
\hline
Sliding-Window & 18.5  &  \textbf{87.3} & 30.5  & 85.4 & 83.4 & 67.0 &  & 13.1 & \textbf{78.6} & 22.1 & 34.2 & 73.2	 & 51.2 \\
SS-TCN~\cite{AbuFarha2019MSTCNMT} &  41.2 & 81.0 &  54.6 & \textbf{86.5} & \textbf{84.2} & 70.3 &  & 28.9 & 64.2 & 38.9 & 61.6 & 77.7 &  53.9\\
MS-TCN~\cite{AbuFarha2019MSTCNMT} &  \textbf{54.7} & 81.0 &  \textbf{65.3} &  86.2 & 83.2 & \textbf{70.5} &  &  \textbf{41.7} & 64.5 & \textbf{49.6} & \textbf{66.1} & \textbf{78.1} &  \textbf{54.0} \\
\hline
\end{tabular}
}
\caption{Comparison of TAS methods. S-\{metric\} refers to segmental, F-\{metric\} to frame-based, and m-\{metric\} to macro. }
\label{tab:tas_main} \vspace{-0.3cm}
\end{table}

\begin{figure}[t]
    \centering
    \begin{minipage}{0.32\textwidth}
        \includegraphics[width=\linewidth]{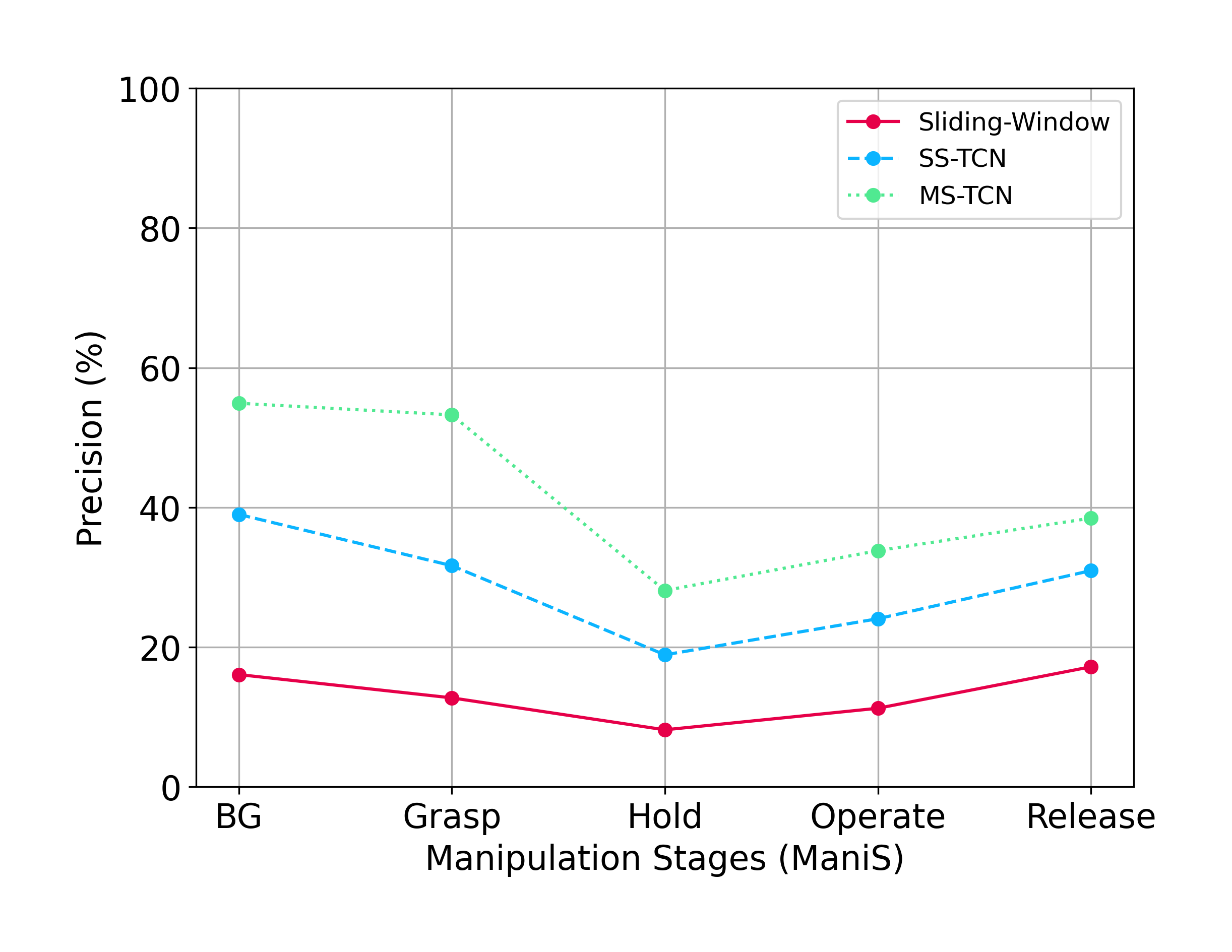}
    \end{minipage}
    \hfill
    \begin{minipage}{0.32\textwidth}
        \includegraphics[width=\linewidth]{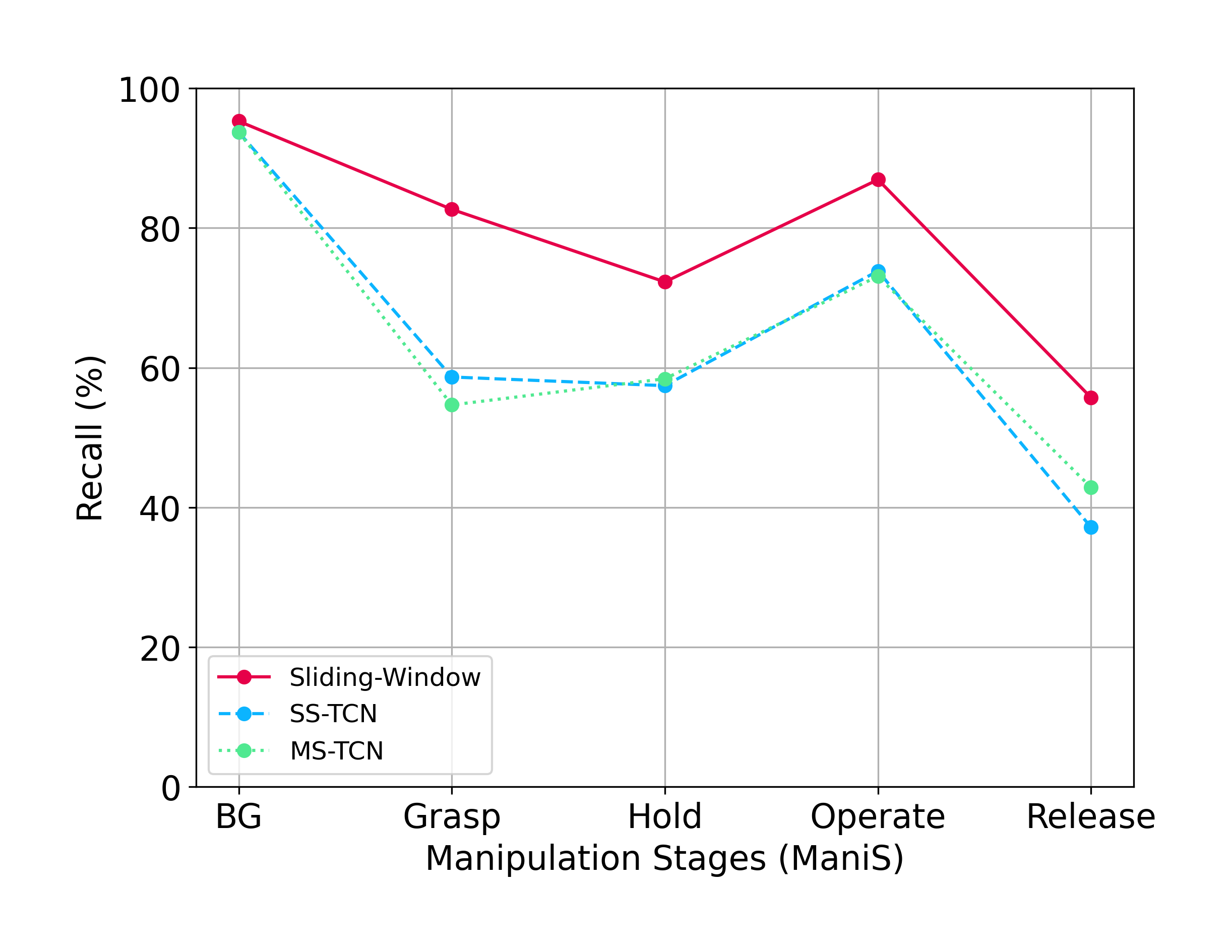}
    \end{minipage}
    \hfill
    \begin{minipage}{0.32\textwidth}
        \includegraphics[width=\linewidth]{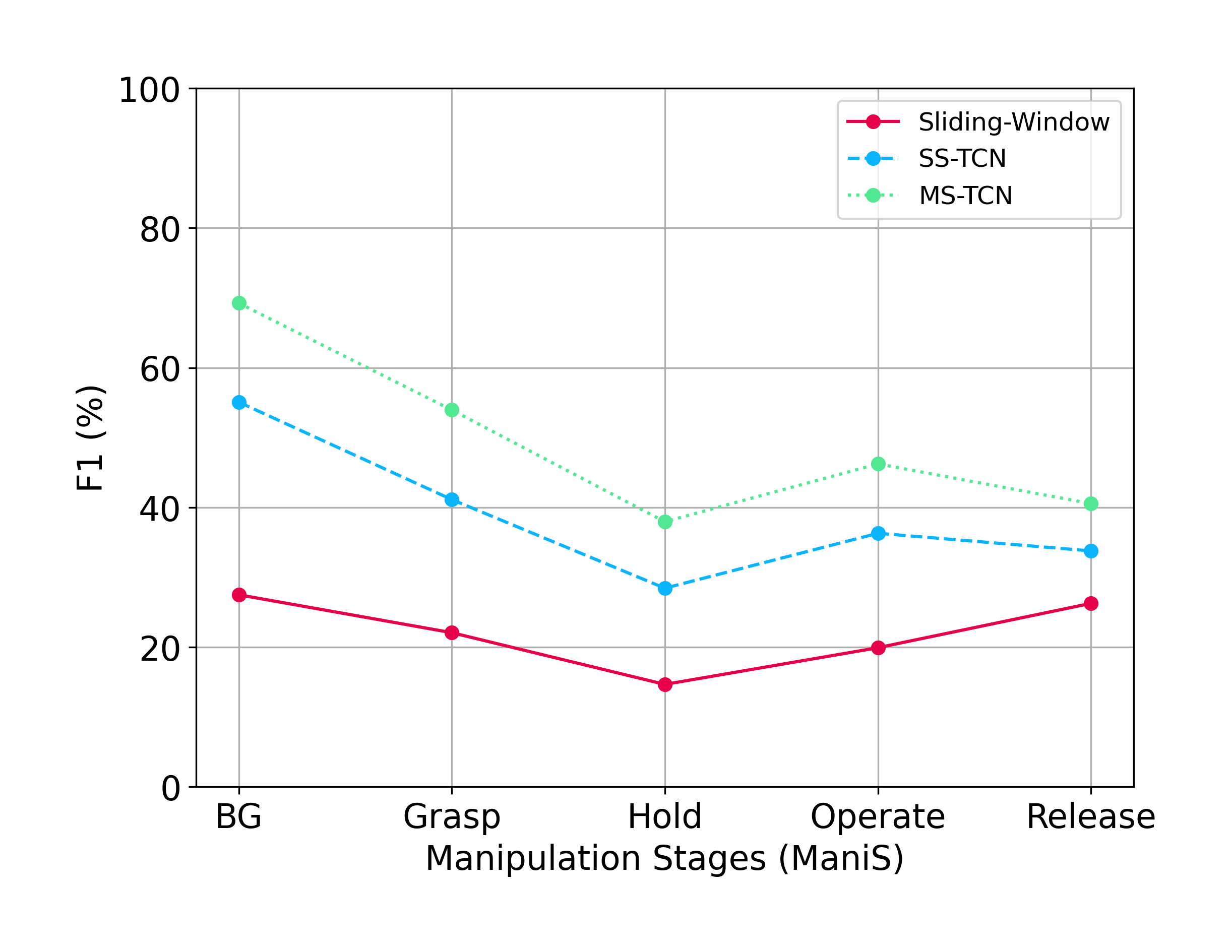}
    \end{minipage}
    \vspace{-3mm}
    \caption{Class-wise segmental: Precision, Recall, and F1.}
    \label{fig:metrics_comparison_seg} \vspace{-0.6cm}
\end{figure}

\begin{figure}[t]
\centering
\includegraphics[width=0.99\textwidth]{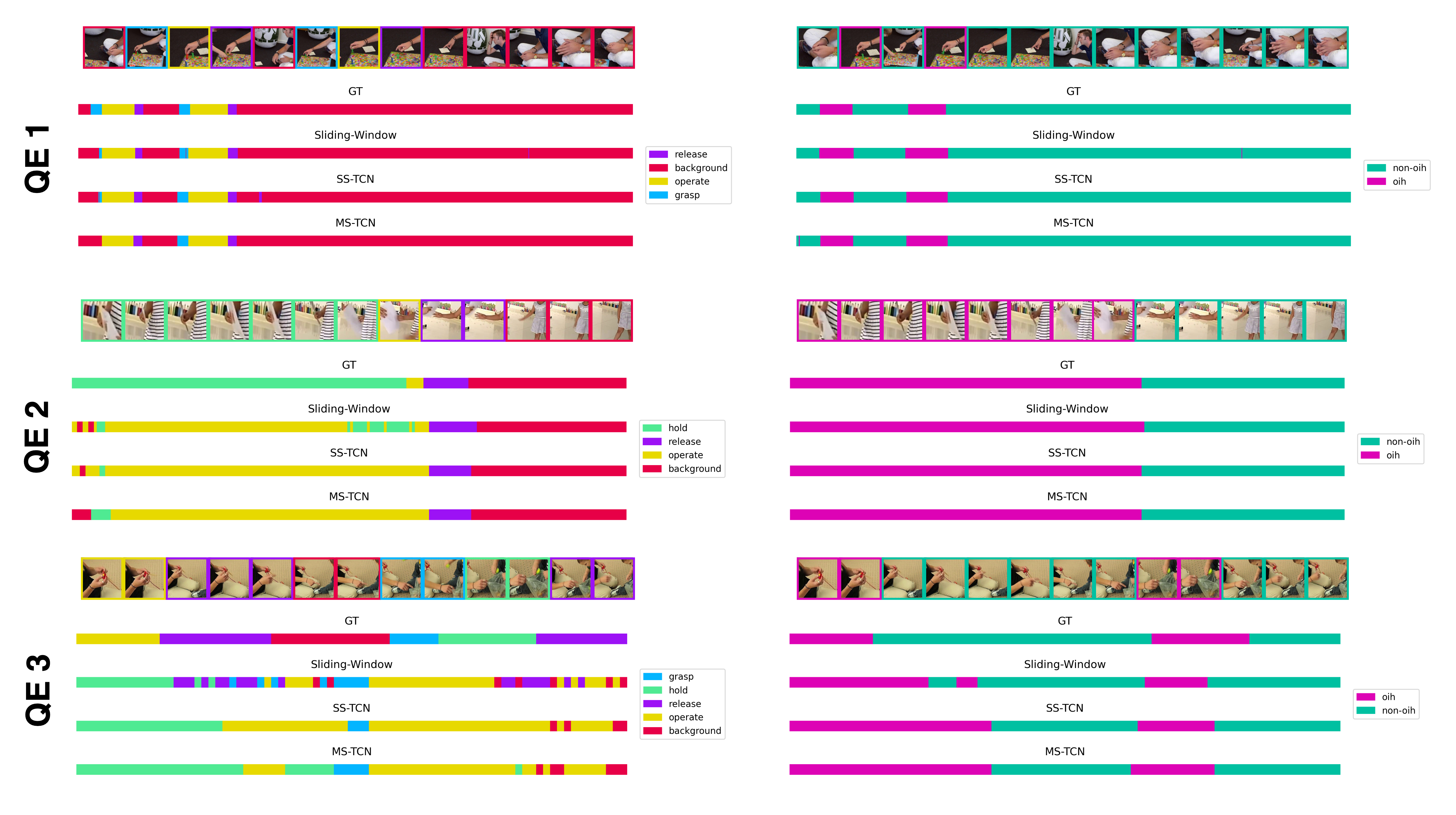}
\vspace{-2mm}

\caption{Qualitative results of predicted segments for hand interaction sequences in ChildPlay-Hand. The left figure corresponds to the ManiS task, and the right figure to the OiH task. Hand-region frames for each segment are highlighted using the same color as their label, representing the key moments within each segment. Note that the hand regions are cropped using the procedure discussed in Sec.~\ref{sec:action_rec}.}
\label{fig:segments} \vspace{-0.6cm}
\end{figure}

\vspace{-1mm}
\subsection{Segmentation Results}
\label{sec:seg_results}

\vspace{-1mm}

\mypartitle{OiH Task:}
Tab.~\ref{tab:tas_main} reports the segmentation results obtained with the methods discussed in Sec.~\ref{sec:tas_methods}. In terms of segmental F1 (S-F1), both SS-TCN and MS-TCN outperform the sliding-window approach, which is expected since the sliding-window method cannot leverage long-range temporal context. 
These methods improve segmental precision (S-Prec.) by removing spurious short term predictions  which strongly affect this metric.
Additionally, they also slightly improve frame-based metrics. 
When comparing SS-TCN and MS-TCN, MS-TCN further improves the segmental metrics while maintaining the same frame accuracy. 

\mypartitle{ManiS Task:} 
Here, MS-TCN also outperforms the other models largely due to the improvement on the precision metric as in the OiH task, at the cost of only a slight degradation of the recall (esp. for \texttt{grasp}). 
We can also notice that the frame-based metrics improves, showing that some regularization helps here as well. 
In general, these results highlight that multiple stages are are beneficial for handling the task's complexity, unlike in OiH where more stages did not improve frame-based metrics. 
Fig.~\ref{fig:metrics_comparison_seg} shows that  MS-TCN consistently achieves the highest F1 scores across all classes.

\mypartitle{Qualitative Examples:}
Fig.~\ref{fig:segments} shows qualitative examples (QEs) of predicted segments. In QE1, all methods perform well on both tasks, although MS-TCN fails to predict the first \texttt{grasp} segment, likely due to over-smoothing. In QE2, while all methods perform well on the OiH task, they struggle to accurately predict the \texttt{hold} segment in the ManiS task. This is due to slight hand motion made by the person while passively holding the object, leading the methods to predict \texttt{operate} instead. In this case, the hand-region alone does not provide enough context to determine whether the person is intentionally moving the object. In QE3, the predictions are of lower quality. For example, in the ManiS task, \texttt{background} is incorrectly predicted as \texttt{grasp}, due to confusion between the motions of pointing and grasping. Also, the last \texttt{release} mostly is predicted as \texttt{operate} while in this stage there is no object in hand. In the OiH task, all methods struggle to accurately predict the boundaries of the OiH state, especially when the hands are close to the object during the \texttt{release}. 

These observations again highlight the difficulty of modeling hands in action in the wild, where hand movements can extend beyond object manipulation (e.g., pointing), objects come in varying shapes and sizes, and multiple people may interact with the same object or be in close proximity, making accurate prediction of hand manipulations challenging.
\vspace{-4mm}
\section{Conclusion}
\vspace{-4mm}
In this work, we introduced ChildPlay-Hand, a novel dataset of hand manipulations in the wild. This dataset is unique for its rich annotations and scenes featuring multiple people interacting naturally with objects and each other in uncontrolled settings. These annotations enable the use of ChildPlay-Hand for a variety of tasks and protocols, including spatio-temporal action localization, pre-segmented action recognition, and human-object interaction. Future research can also leverage the accompanying gaze labels to explore the coordination between manipulations and visual attention. In this work, we proposed two specific tasks under recognition and temporal segmentation protocols: object in hand (OiH) and manipulation stages (ManiS). We benchmarked various spatio-temporal models with varied modalities and input types as well as segmentation models, on these tasks. 
Our findings indicate that ChildPlay-Hand can serve as a challenging benchmark for understanding hands in action from a third-person view in uncontrolled settings.

\mypartitle{Acknowledgement.}
This research was supported by the AI4Autism project (digital phenotyping of autism spectrum disorders in children, grant agreement number CRSII5\_202235 / 1) of the Sinergia interdisciplinary program of the SNSF.

\bibliographystyle{splncs04}
\bibliography{main}

\clearpage
\begin{center}
    \Large
    \textbf{ChildPlay-Hand: A Dataset of Hand Manipulations in the Wild}\\
    \vspace{0.5em}Supplementary Material \\
\end{center}

\vspace{-4mm}
\section{More Statistics from the ChildPlay-Hand Dataset}
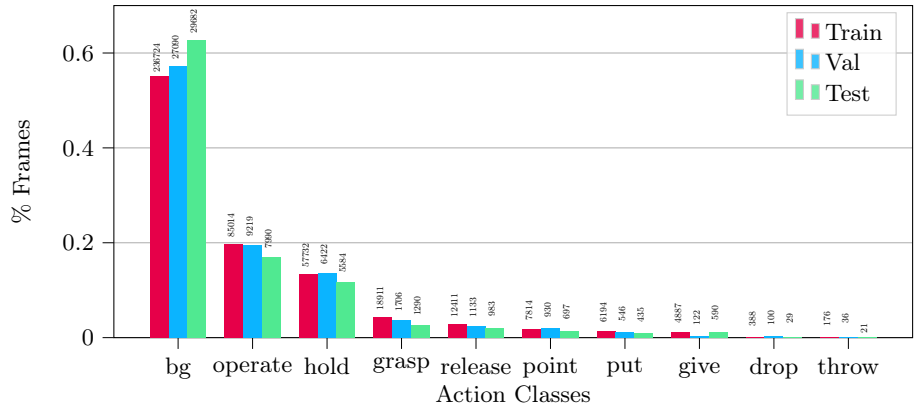
\begin{figure}
\centering

\begin{subfigure}[b]{\textwidth}
\begin{tikzpicture}

\definecolor{crimson230073}{RGB}{230,0,73}
\definecolor{darkgray176}{RGB}{176,176,176}
\definecolor{deepskyblue11180255}{RGB}{11,180,255}
\definecolor{lightgray204}{RGB}{204,204,204}
\definecolor{mediumaquamarine80233145}{RGB}{80,233,145}

\begin{axis}[
    width=\textwidth,
    height=6cm,
    legend cell align={left},
    legend style={fill opacity=0.8, draw opacity=1, text opacity=1, draw=lightgray204},
    tick align=outside,
    tick pos=left,
    x grid style={darkgray176},
    xlabel={Action Classes},
    xmin=-0.8625, xmax=9.8625,
    xtick style={color=black},
    xtick={0,1,2,3,4,5,6,7,8,9},
    xticklabels={bg,operate,hold,grasp,release,point,put,give,drop,throw},
    y grid style={darkgray176},
    ylabel={\% Frames},
    ymajorgrids,
    ymin=0, ymax=0.7,
    ytick style={color=black}
]
\draw[draw=none,fill=crimson230073] (axis cs:-0.375,0) rectangle (axis cs:-0.125,0.550198488327771);
\addlegendimage{ybar,ybar legend,draw=none,fill=crimson230073}
\addlegendentry{Train}

\draw[draw=none,fill=crimson230073] (axis cs:0.625,0) rectangle (axis cs:0.875,0.197591179122933);
\draw[draw=none,fill=crimson230073] (axis cs:1.625,0) rectangle (axis cs:1.875,0.134184152543161);
\draw[draw=none,fill=crimson230073] (axis cs:2.625,0) rectangle (axis cs:2.875,0.0439533110828073);
\draw[draw=none,fill=crimson230073] (axis cs:3.625,0) rectangle (axis cs:3.875,0.0288458856670045);
\draw[draw=none,fill=crimson230073] (axis cs:4.625,0) rectangle (axis cs:4.875,0.0181614495690897);
\draw[draw=none,fill=crimson230073] (axis cs:5.625,0) rectangle (axis cs:5.875,0.0143962143116127);
\draw[draw=none,fill=crimson230073] (axis cs:6.625,0) rectangle (axis cs:6.875,0.011358459693389);
\draw[draw=none,fill=crimson230073] (axis cs:7.625,0) rectangle (axis cs:7.875,0.000901797086358692);
\draw[draw=none,fill=crimson230073] (axis cs:8.625,0) rectangle (axis cs:8.875,0.000409062595874046);
\draw[draw=none,fill=deepskyblue11180255] (axis cs:-0.125,0) rectangle (axis cs:0.125,0.572654631547795);
\addlegendimage{ybar,ybar legend,draw=none,fill=deepskyblue11180255}
\addlegendentry{Val}

\draw[draw=none,fill=deepskyblue11180255] (axis cs:0.875,0) rectangle (axis cs:1.125,0.194880142053862);
\draw[draw=none,fill=deepskyblue11180255] (axis cs:1.875,0) rectangle (axis cs:2.125,0.135775588720247);
\draw[draw=none,fill=deepskyblue11180255] (axis cs:2.875,0) rectangle (axis cs:3.125,0.0360630786792373);
\draw[draw=none,fill=deepskyblue11180255] (axis cs:3.875,0) rectangle (axis cs:4.125,0.0239504502600093);
\draw[draw=none,fill=deepskyblue11180255] (axis cs:4.875,0) rectangle (axis cs:5.125,0.0196592398427261);
\draw[draw=none,fill=deepskyblue11180255] (axis cs:5.875,0) rectangle (axis cs:6.125,0.0115418762947618);
\draw[draw=none,fill=deepskyblue11180255] (axis cs:6.875,0) rectangle (axis cs:7.125,0.00260009301145732);
\draw[draw=none,fill=deepskyblue11180255] (axis cs:7.875,0) rectangle (axis cs:8.125,0.00211389675728237);
\draw[draw=none,fill=deepskyblue11180255] (axis cs:8.875,0) rectangle (axis cs:9.125,0.000761002832621655);
\draw[draw=none,fill=mediumaquamarine80233145] (axis cs:0.125,0) rectangle (axis cs:0.375,0.627456737921854);
\addlegendimage{ybar,ybar legend,draw=none,fill=mediumaquamarine80233145}
\addlegendentry{Test}

\draw[draw=none,fill=mediumaquamarine80233145] (axis cs:1.125,0) rectangle (axis cs:1.375,0.168917860778326);
\draw[draw=none,fill=mediumaquamarine80233145] (axis cs:2.125,0) rectangle (axis cs:2.375,0.118040873854827);
\draw[draw=none,fill=mediumaquamarine80233145] (axis cs:3.125,0) rectangle (axis cs:3.375,0.027288387753504);
\draw[draw=none,fill=mediumaquamarine80233145] (axis cs:4.125,0) rectangle (axis cs:4.375,0.0207892882311487);
\draw[draw=none,fill=mediumaquamarine80233145] (axis cs:5.125,0) rectangle (axis cs:5.375,0.0147404275311252);
\draw[draw=none,fill=mediumaquamarine80233145] (axis cs:6.125,0) rectangle (axis cs:6.375,0.00920053245634641);
\draw[draw=none,fill=mediumaquamarine80233145] (axis cs:7.125,0) rectangle (axis cs:7.375,0.012489233419466);
\draw[draw=none,fill=mediumaquamarine80233145] (axis cs:8.125,0) rectangle (axis cs:8.375,0.000626419231070394);
\draw[draw=none,fill=mediumaquamarine80233145] (axis cs:9.125,0) rectangle (axis cs:9.375,0.000450238822331846);
\draw (axis cs:-0.2,0.590198488327771) node[
  scale=0.4,
  anchor=south,
  text=black,
  rotate=90.0
]{236724};
\draw (axis cs:0.80,0.237591179122933) node[
  scale=0.4,
  anchor=south,
  text=black,
  rotate=90.0
]{85014};
\draw (axis cs:1.8,0.174184152543161) node[
  scale=0.4,
  anchor=south,
  text=black,
  rotate=90.0
]{57732};
\draw (axis cs:2.8,0.0839533110828073) node[
  scale=0.4,
  anchor=south,
  text=black,
  rotate=90.0
]{18911};
\draw (axis cs:3.8,0.0688458856670045) node[
  scale=0.4,
  anchor=south,
  text=black,
  rotate=90.0
]{12411};
\draw (axis cs:4.8,0.0581614495690897) node[
  scale=0.4,
  anchor=south,
  text=black,
  rotate=90.0
]{7814};
\draw (axis cs:5.8,0.0543962143116127) node[
  scale=0.4,
  anchor=south,
  text=black,
  rotate=90.0
]{6194};
\draw (axis cs:6.8,0.051358459693389) node[
  scale=0.4,
  anchor=south,
  text=black,
  rotate=90.0
]{4887};
\draw (axis cs:7.8,0.0409017970863587) node[
  scale=0.4,
  anchor=south,
  text=black,
  rotate=90.0
]{388};
\draw (axis cs:8.8,0.040409062595874) node[
  scale=0.4,
  anchor=south,
  text=black,
  rotate=90.0
]{176};
\draw (axis cs:0.05,0.612654631547795) node[
  scale=0.4,
  anchor=south,
  text=black,
  rotate=90.0
]{27090};
\draw (axis cs:1.05,0.234880142053862) node[
  scale=0.4,
  anchor=south,
  text=black,
  rotate=90.0
]{9219};
\draw (axis cs:2.05,0.175775588720247) node[
  scale=0.4,
  anchor=south,
  text=black,
  rotate=90.0
]{6422};
\draw (axis cs:3.05,0.0760630786792373) node[
  scale=0.4,
  anchor=south,
  text=black,
  rotate=90.0
]{1706};
\draw (axis cs:4.05,0.0639504502600093) node[
  scale=0.4,
  anchor=south,
  text=black,
  rotate=90.0
]{1133};
\draw (axis cs:5.05,0.0596592398427261) node[
  scale=0.4,
  anchor=south,
  text=black,
  rotate=90.0
]{930};
\draw (axis cs:6.05,0.0515418762947618) node[
  scale=0.4,
  anchor=south,
  text=black,
  rotate=90.0
]{546};
\draw (axis cs:7.05,0.0426000930114573) node[
  scale=0.4,
  anchor=south,
  text=black,
  rotate=90.0
]{122};
\draw (axis cs:8.05,0.0421138967572824) node[
  scale=0.4,
  anchor=south,
  text=black,
  rotate=90.0
]{100};
\draw (axis cs:9.05,0.0407610028326217) node[
  scale=0.4,
  anchor=south,
  text=black,
  rotate=90.0
]{36};
\draw (axis cs:0.3,0.667456737921854) node[
  scale=0.4,
  anchor=south,
  text=black,
  rotate=90.0
]{29682};
\draw (axis cs:1.3,0.208917860778326) node[
  scale=0.4,
  anchor=south,
  text=black,
  rotate=90.0
]{7990};
\draw (axis cs:2.3,0.158040873854827) node[
  scale=0.4,
  anchor=south,
  text=black,
  rotate=90.0
]{5584};
\draw (axis cs:3.3,0.067288387753504) node[
  scale=0.4,
  anchor=south,
  text=black,
  rotate=90.0
]{1290};
\draw (axis cs:4.3,0.0607892882311487) node[
  scale=0.4,
  anchor=south,
  text=black,
  rotate=90.0
]{983};
\draw (axis cs:5.3,0.0547404275311252) node[
  scale=0.4,
  anchor=south,
  text=black,
  rotate=90.0
]{697};
\draw (axis cs:6.3,0.0492005324563464) node[
  scale=0.4,
  anchor=south,
  text=black,
  rotate=90.0
]{435};
\draw (axis cs:7.3,0.052489233419466) node[
  scale=0.4,
  anchor=south,
  text=black,
  rotate=90.0
]{590};
\draw (axis cs:8.3,0.0406264192310704) node[
  scale=0.4,
  anchor=south,
  text=black,
  rotate=90.0
]{29};
\draw (axis cs:9.3,0.0204502388223318) node[
  scale=0.4,
  anchor=south,
  text=black,
  rotate=90.0
]{21};
\end{axis}

\end{tikzpicture}
\end{subfigure}

\caption{Distribution of annotations per train/val/test splits. We show the distribution in frames (on top of each bar) and in percentage (y-axis).}
\label{fig:childplay-hand-splits-stats} \vspace{-6mm}
\end{figure}

\vspace{-4mm}
\sam{The dataset partitioning for training, validation and testing follows the same splits introduced in \cite{tafasca2023childplay}, as we found them to be well balanced for the new annotation task already. Specifically, the class distributions across these splits were fairly similar, as can be shown in Figure \ref{fig:childplay-hand-splits-stats}. However, it is worth noting that the classes with a low frequency evidently feature even fewer instances after splitting (\eg the \texttt{throw} action has only $36$ and $21$ frames in the validation and test sets, respectively).}

\vspace{-4mm}
\section{Implementation Details}
\label{sec:sup_impl_details}
\vspace{-3mm}
\mypartitle{Training instances: }To create training instances for recognition methods, we sample key frames to serve as the target class for classification and take frames around them to create a short segment. We use every frame in the training set as the middle frame for classification, except for the background class, where we sample every third frame. Frames where the person is occluded are excluded from both training and evaluation. 

\mypartitle{Training details of recognition: }When full-body information is used as input, an instance is a person, so we use two classification heads, one for each hand (left and right). During training, we apply horizontal flip as the only augmentation, flipping the ground-truth hand labels accordingly. However, when the hand is used as the input, an instance is a single hand of a person, so we use only one classification head and do not flip the labels when horizontal flip is applied. The classification head consists of two MLPs with a ReLU activation in between, which map the temporally mean-pooled features to one of the C categories. We fine-tune all recognition networks for 15 epochs. For PoseConv3D and RGBPoseConv3D models, we use SGD optimizer and set the learning rate of the backbone to 0.0018 and the MLP heads to 0.018, with linear decay at epochs 9 and 13. For Hiera models, we use Hiera-B version and finetune it using the AdamW~\cite{Loshchilov2017DecoupledWD} optimizer with a learning rate of 2e-5 for the backbone and 2e-4 for the MLP heads, applying cosine decay as the scheduler.

\mypartitle{Training details of segmentation: }We train MS-TCN for 50 epochs using the Adam~\cite{Kingma2014AdamAM} optimizer with a learning rate of 0.0005, selecting the best checkpoint based on the validation set. The number of layers is set to 10 for both SS-TCN and MS-TCN, with the number of stages set to 4 for MS-TCN. 

\begin{figure}[t]
\centering
\includegraphics[width=0.99\textwidth]{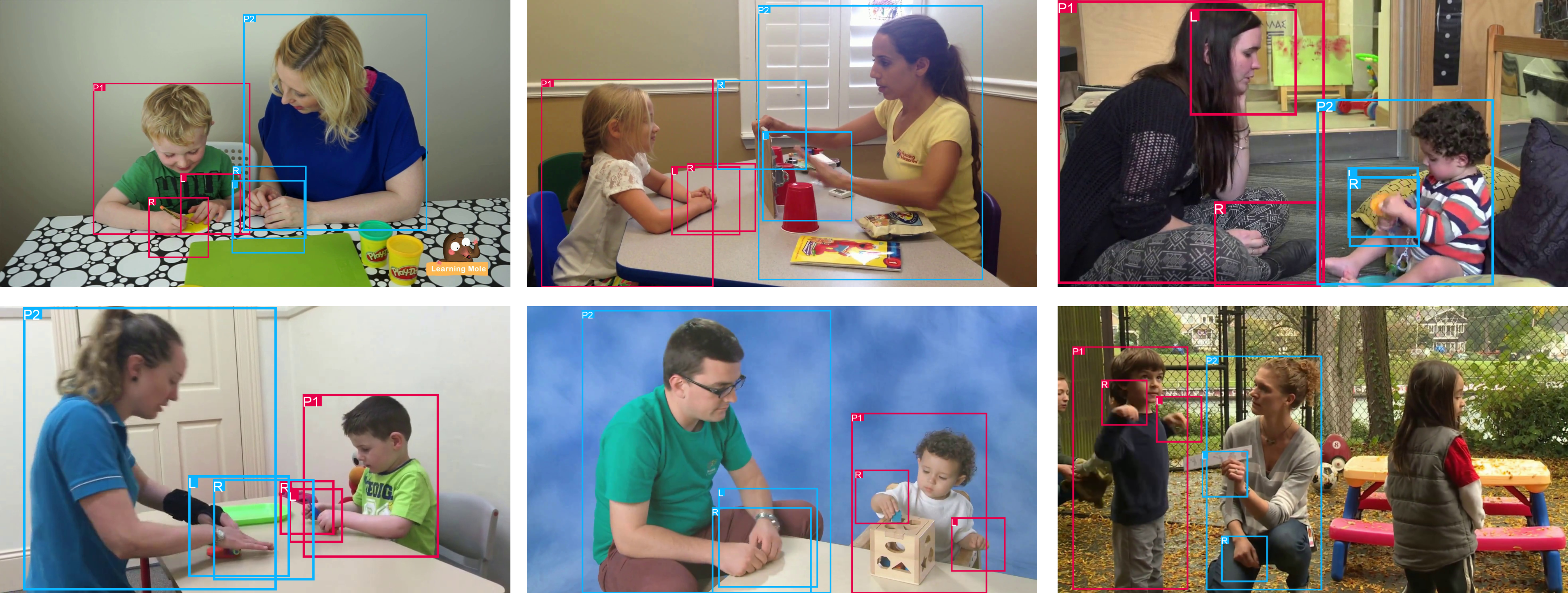}
\vspace{-0.3cm}
\caption{Examples of the ground-truth body and pseudo-hand bounding boxes in ChildPlay-Hand. These bounding boxes are used in the cropping strategy described in Sec.~\textcolor{red}{4.2} of the main submission to create input volumes for the networks.}
\label{fig:hand_crop} \vspace{-0.6cm}
\end{figure}

\vspace{-3mm}
\section{Generating Pseudo-Hand Bounding Boxes }
\label{sec:sup_pseudo_hand_box}
\vspace{-3mm}
We generate pseudo-hand bounding boxes as follows: (1) we define the center of the box at the wrist keypoint, extended by 50\% of the elbow-to-wrist limb, and (2) set the box size to 40\% of the smallest side of the body bounding box. Note, while a hand detector could typically be used, detecting and associating hands with body bounding boxes in our unconstrained setting is challenging. Using hand keypoints from a whole-body pose estimator is another option, but these keypoints can be noisy due to self/object occlusions, leading to inaccurate hand bounding boxes. Instead, we find that the wrist and elbow keypoints from 2D body pose estimators like HRNet~\cite{Wang2019DeepHR} is robust enough for generating pseudo-hand bounding boxes. See illustrations in Figure~\ref{fig:hand_crop}. 

\vspace{-3mm}
\section{On confusion among manipulation categories}
\begin{figure}[t]
    \centering
    \begin{minipage}{0.32\textwidth}
        \includegraphics[width=\linewidth]{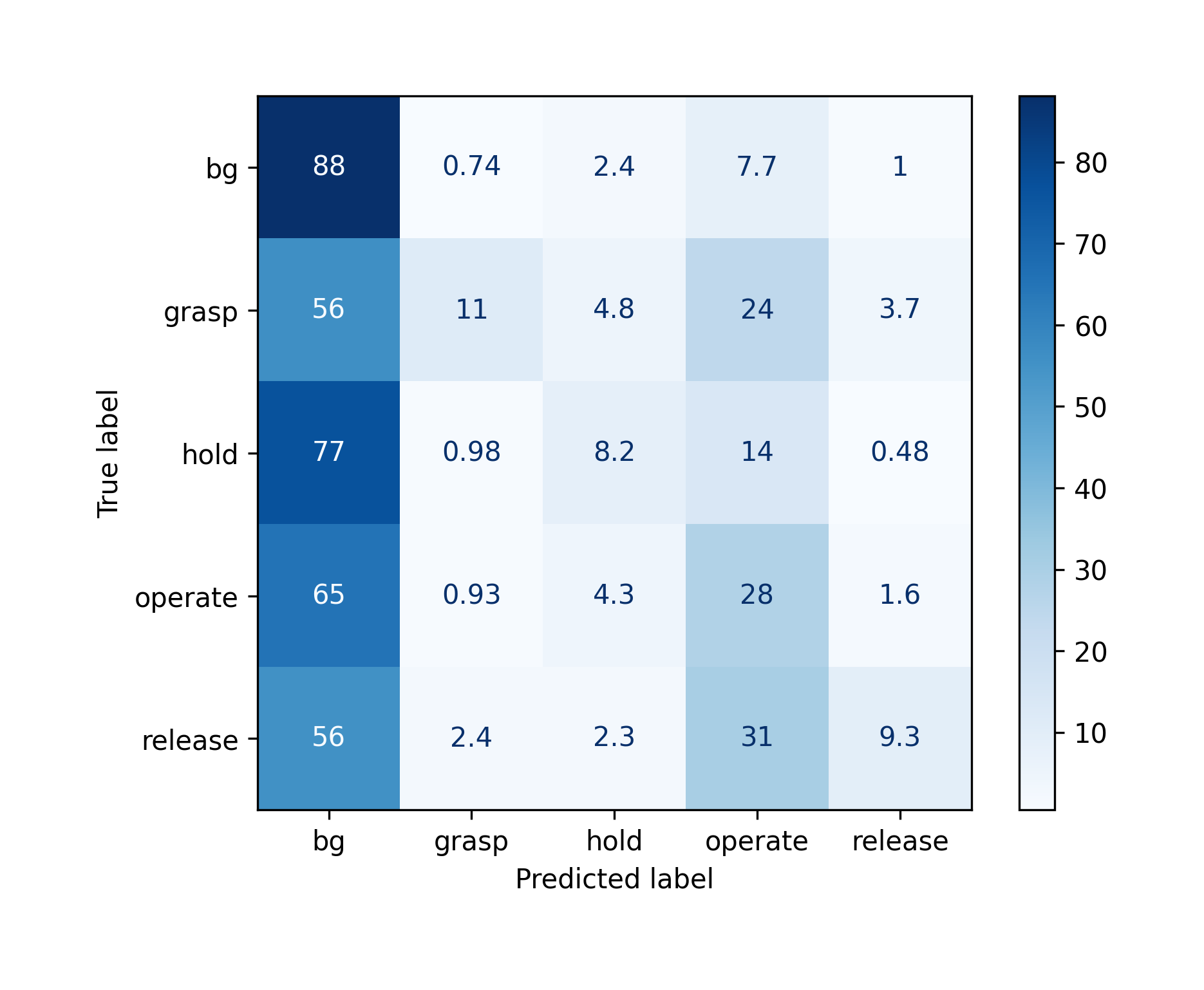}
    \end{minipage}
    \hfill
    \begin{minipage}{0.32\textwidth}
        \includegraphics[width=\linewidth]{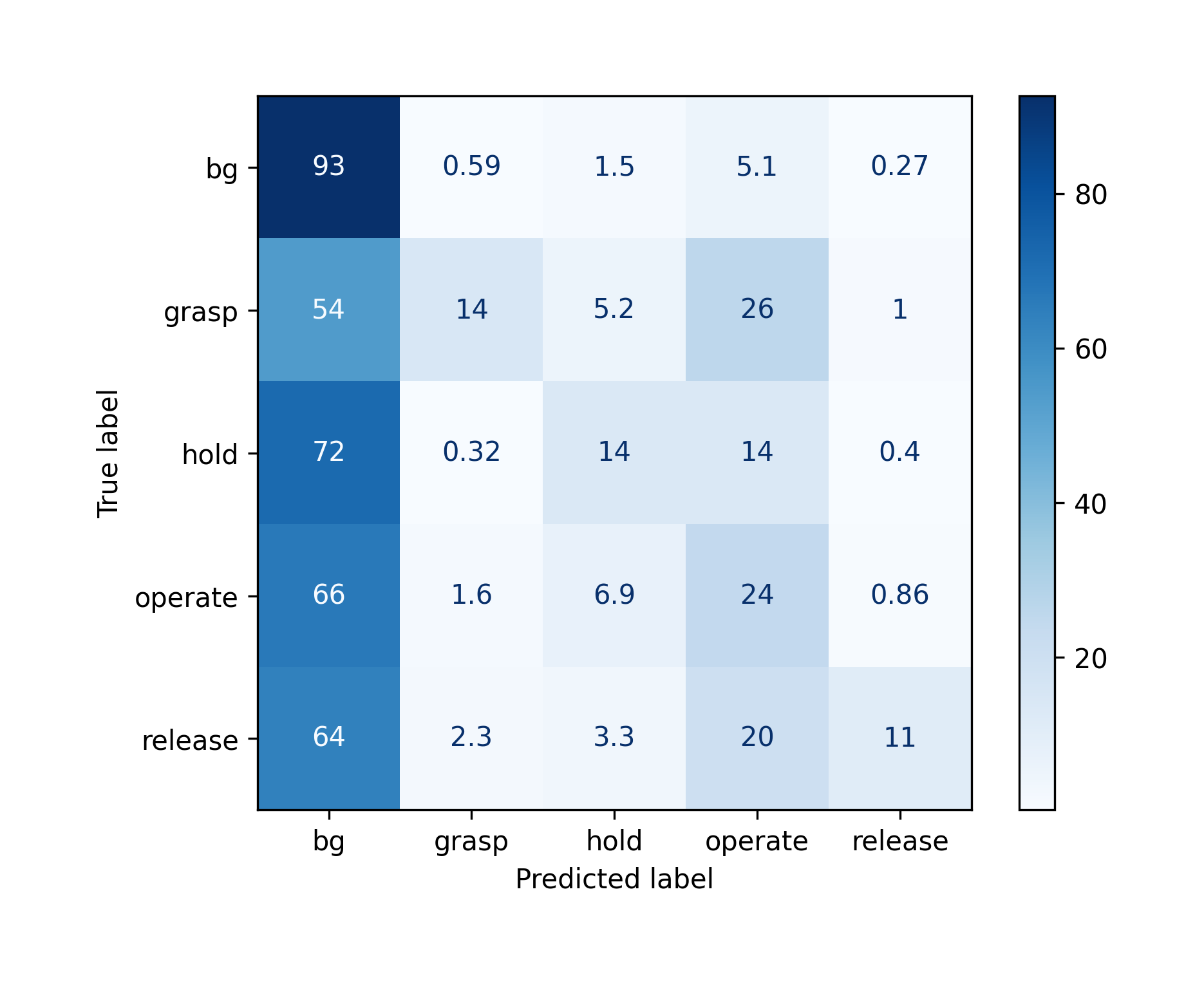}
    \end{minipage}
    \hfill
    \begin{minipage}{0.32\textwidth}
        \includegraphics[width=\linewidth]{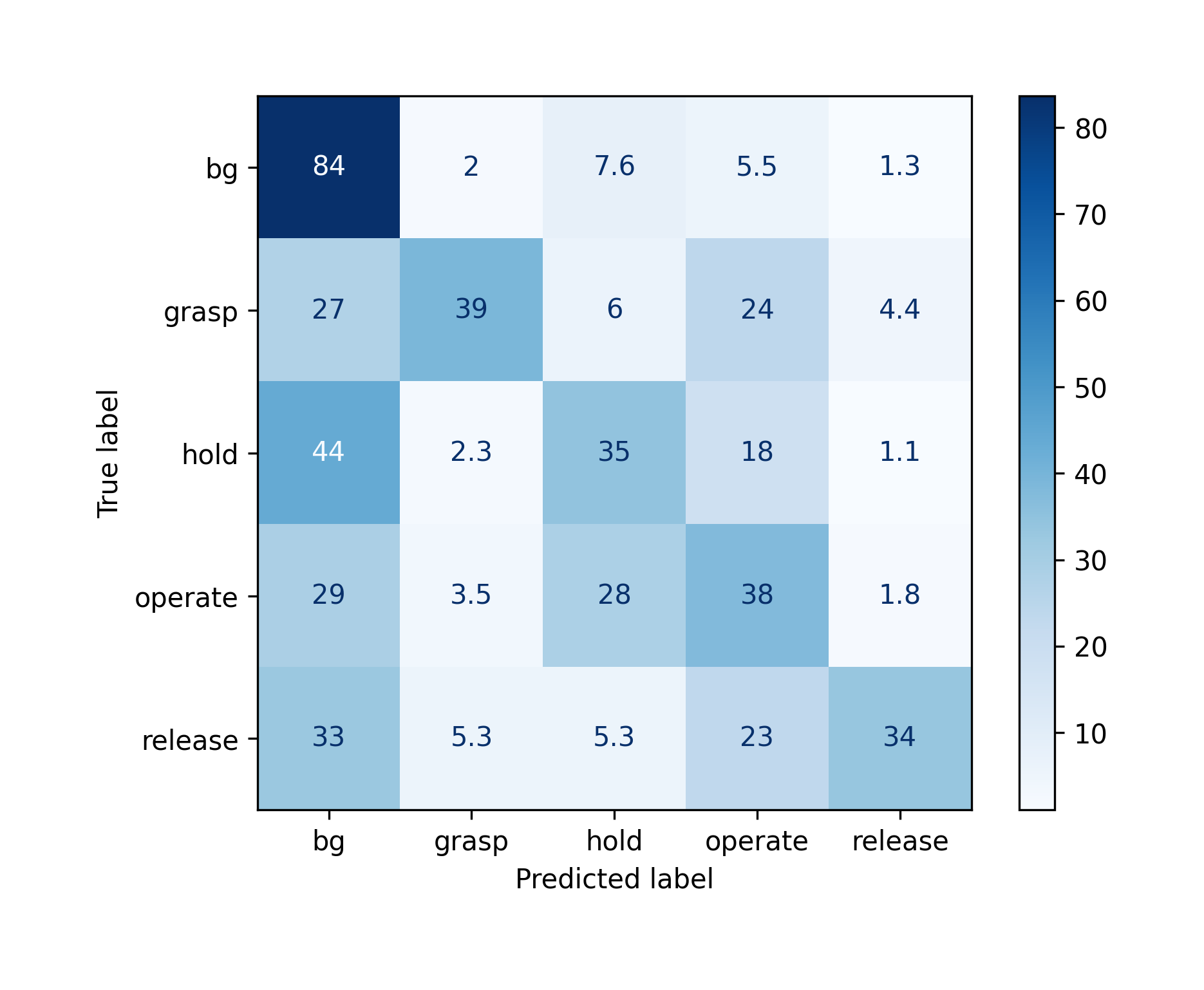}
    \end{minipage}
    \vspace{-3mm}
    \caption{Confusion Matrices of PoseConv3D, RGBPoseConv3D, and Hiera networks with \texttt{body} input.}
    \label{fig:conf_body}
    
    \vspace{2mm}
    
    \begin{minipage}{0.32\textwidth}
        \includegraphics[width=\linewidth]{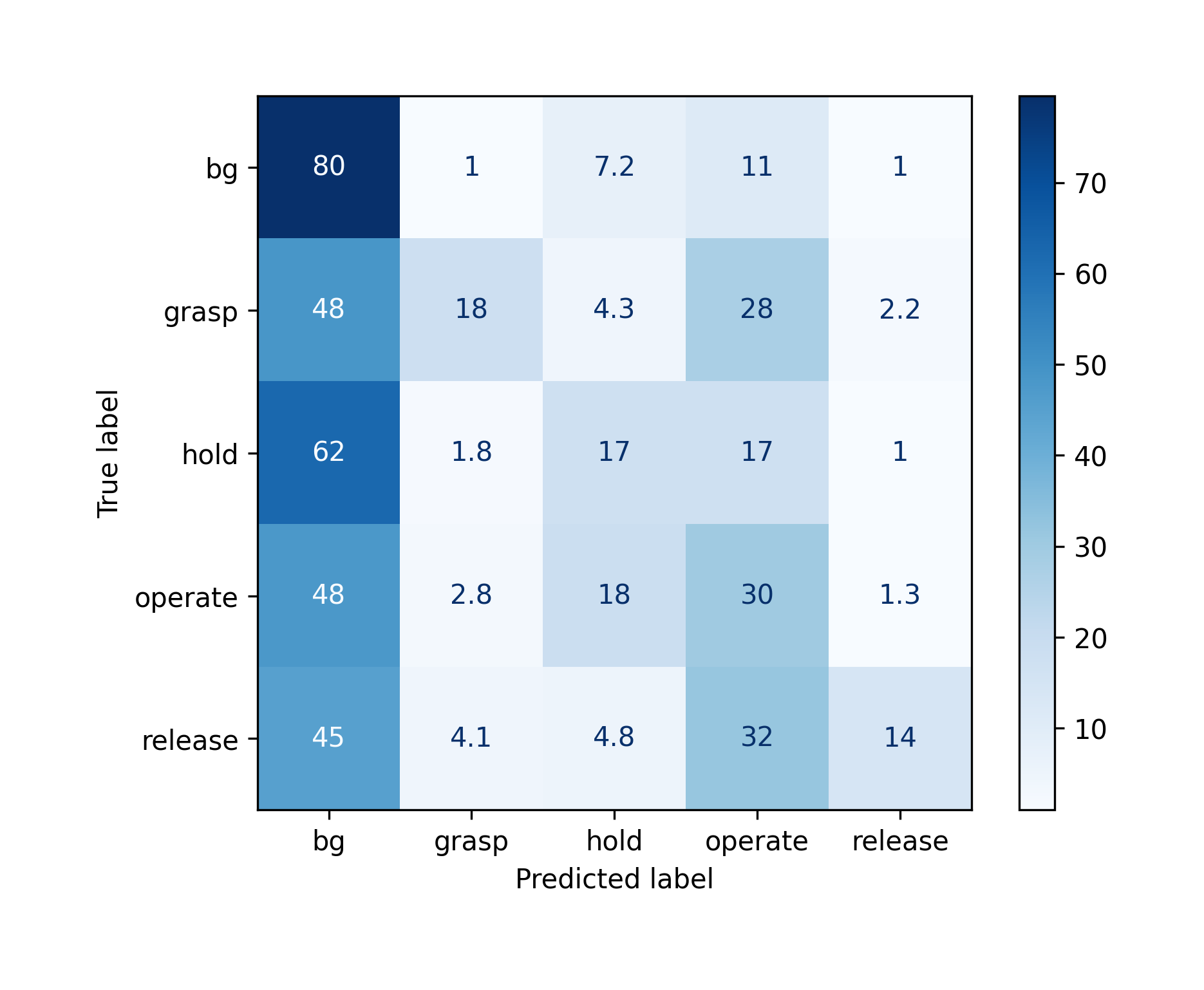}
    \end{minipage}
    \hfill
    \begin{minipage}{0.32\textwidth}
        \includegraphics[width=\linewidth]{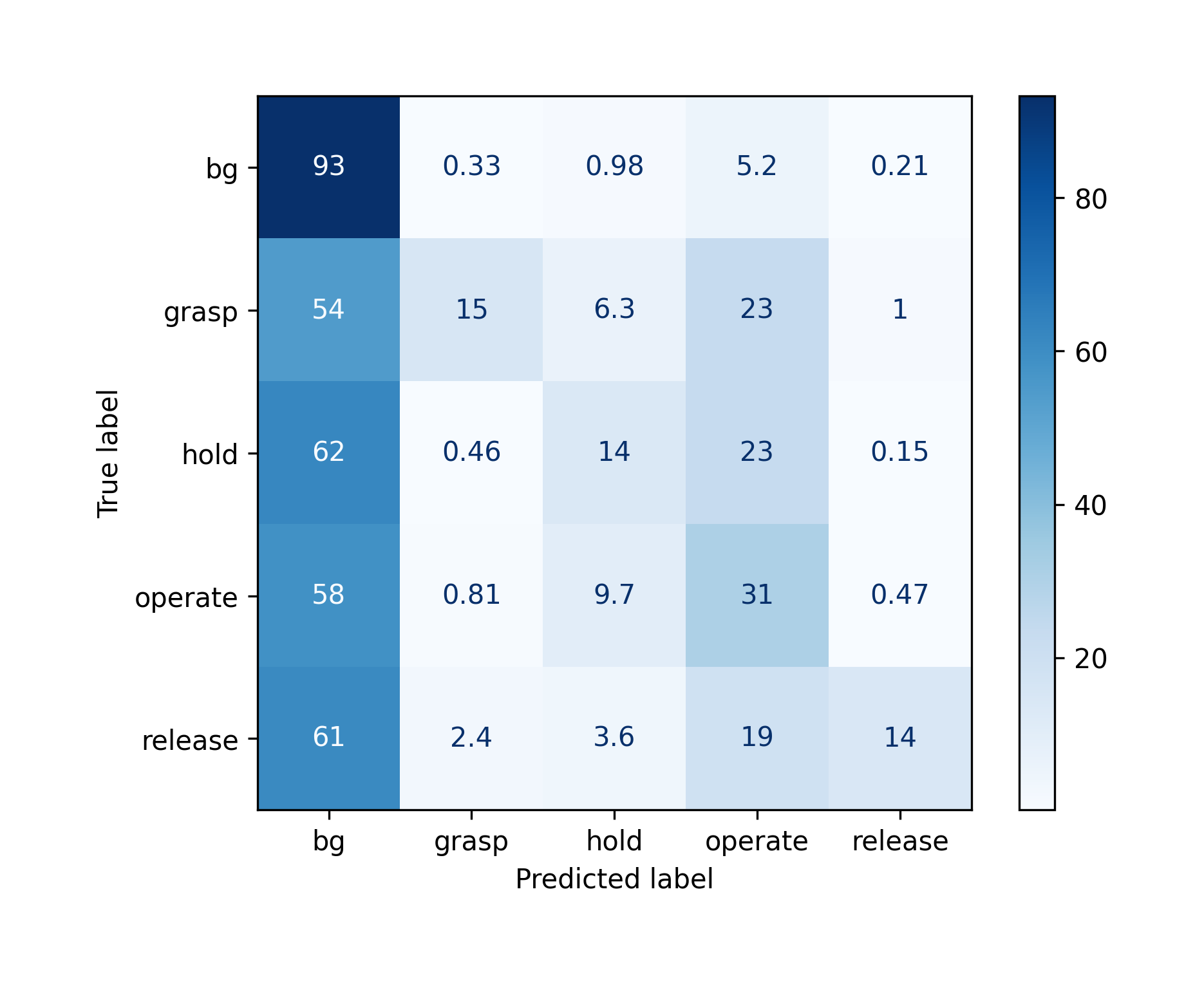}
    \end{minipage}
    \hfill
    \begin{minipage}{0.32\textwidth}
        \includegraphics[width=\linewidth]{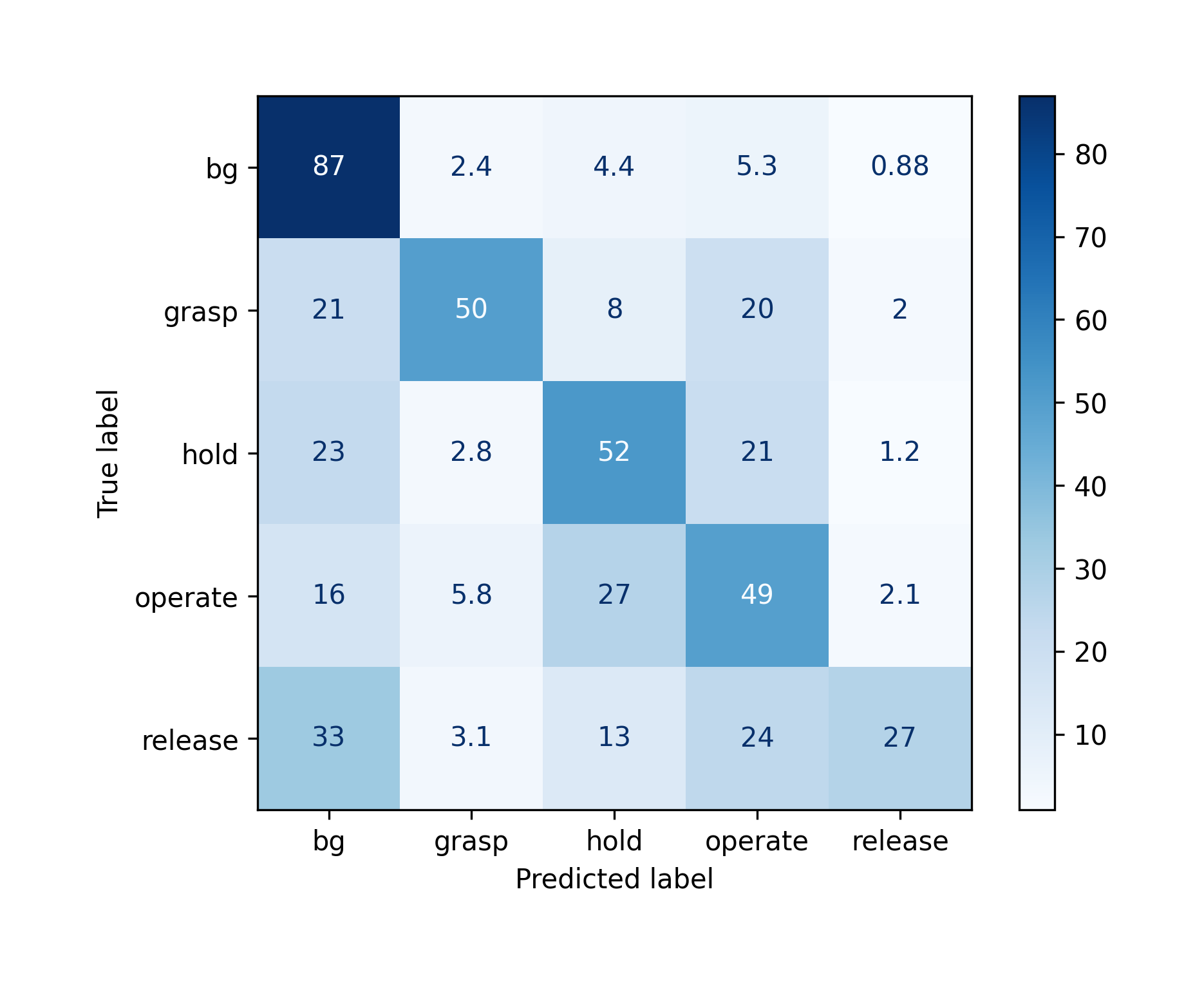}
    \end{minipage}
    \vspace{-3mm}
    \caption{Confusion Matrices of PoseConv3D, RGBPoseConv3D, and Hiera networks with \texttt{hand} input.}
    \label{fig:conf_hand}  \vspace{-5mm}
\end{figure}

\vspace{-3mm}
After looking at the confusion matrices in Fig.~\ref{fig:conf_body} and Fig.~\ref{fig:conf_hand}, we can see a common trend that the major confusion, especially for classes with fewer instances, namely \texttt{grasp} and \texttt{release}, occurs with the majority classes, \texttt{background} and \texttt{operate}. One way to mitigate this is to perform balanced sampling to create balanced mini-batches or apply different weights for the classes at the loss level. Moreover, there is an inherent confusion between \texttt{hold} and \texttt{operate} since these two classes share the same OiH state, but what differentiates them is not only motion but also intention, making it challenging to distinguish between them, even during annotation. Gaze can potentially help in such cases by acting as a cue to infer intention. \vspace{-3mm}

\end{document}